\documentclass[runningheads]{llncs}

\usepackage{eccv}

\usepackage{multirow}
\usepackage{colortbl}   
\usepackage{xcolor}     
\usepackage{makecell}
\usepackage{tabularx}
\usepackage{bm}
\usepackage{stfloats}
\usepackage{placeins}
\usepackage{wrapfig}
\usepackage{enumitem} 
\usepackage[table,dvipsnames]{xcolor} 

\usepackage{tablefootnote}
\usepackage{hyperref}
\usepackage{pifont}
\definecolor{navyblue}{HTML}{0071BC}
\definecolor{mygray}{gray}{0.6}
\newcommand{\cmark}{\color{green}{\ding{51}}}%
\newcommand{\xmark}{\color{red}{\ding{55}}}%
\usepackage{eccvabbrv}

\usepackage{graphicx}
\usepackage{booktabs}
\usepackage{fontawesome5}
\usepackage[accsupp]{axessibility}  

\usepackage{hyperref}

\usepackage{orcidlink}

\begin{document}

\title{Stream3D-VLM: Online 3D Spatial Understanding with Incremental Geometry Priors} 

\titlerunning{Stream3D-VLM}

\author{%
  Hanxun Yu\inst{1,2}\textsuperscript{$\ast$\dag} \and
  Xuan Qu\inst{1,2}\textsuperscript{\dag} \and
  Lei Ke\inst{2} \and
  Boqiang Zhang\inst{2} \and
  Yuxin Wang\inst{2,3} \and
  Jianke Zhu\inst{1,4} \and
  Dong Yu\inst{2}
}

\authorrunning{H.~Yu et al.}
\institute{%
  \textsuperscript{1}Zhejiang University,\enspace
  \textsuperscript{2}Tencent Hunyuan,\enspace
  \textsuperscript{3}HKUST,\enspace
  \textsuperscript{4}Shenzhen Loop Area Institute
  \\ \vspace{10pt}
  \faGithub\ \textbf{Project Page:} \href{https://stream3d-vlm.github.io/}{https://stream3d-vlm.github.io/}
}

\renewcommand{\thefootnote}{$\ast$}
\footnotetext[1]{Work done during an internship at Tencent Hunyuan.}
\renewcommand{\thefootnote}{\dag}
\footnotetext[2]{Equal contribution.}

\maketitle

\begin{abstract}
  Despite advances in 3D scene understanding, existing 3D Large Multimodal Models operate in offline settings, requiring complete scene observations or predefined video clips. In this paper, we present an online 3D vision-language model that enables real-time spatial understanding from streaming video. Our approach adopts an autoregressive streaming control modeling based on the LLM's next-token prediction objective to learn when to respond, and employs a lightweight Visual–Spatial Feature Integration (VSFI) module to incrementally inject temporally aligned geometry priors into the visual stream. To alleviate long-context decoding overhead, we propose a plug-and-play Geometry-Adaptive Voxel Compression (GAVC) module for efficient visual token compression. To address the scarcity of streaming 3D–language data, we further develop a scalable data generation pipeline that curates over 1M online spatio-temporal 3D QA pairs and establishes a comprehensive benchmark spanning 29 tasks. Extensive experiments show that our approach significantly outperforms both proprietary and open-source models across online and offline 3D spatial understanding, reasoning, and grounding tasks.
\end{abstract}

\section{Introduction}
\label{sec:intro}
\begin{figure}[t]
  \centering
  \includegraphics[width=\linewidth]{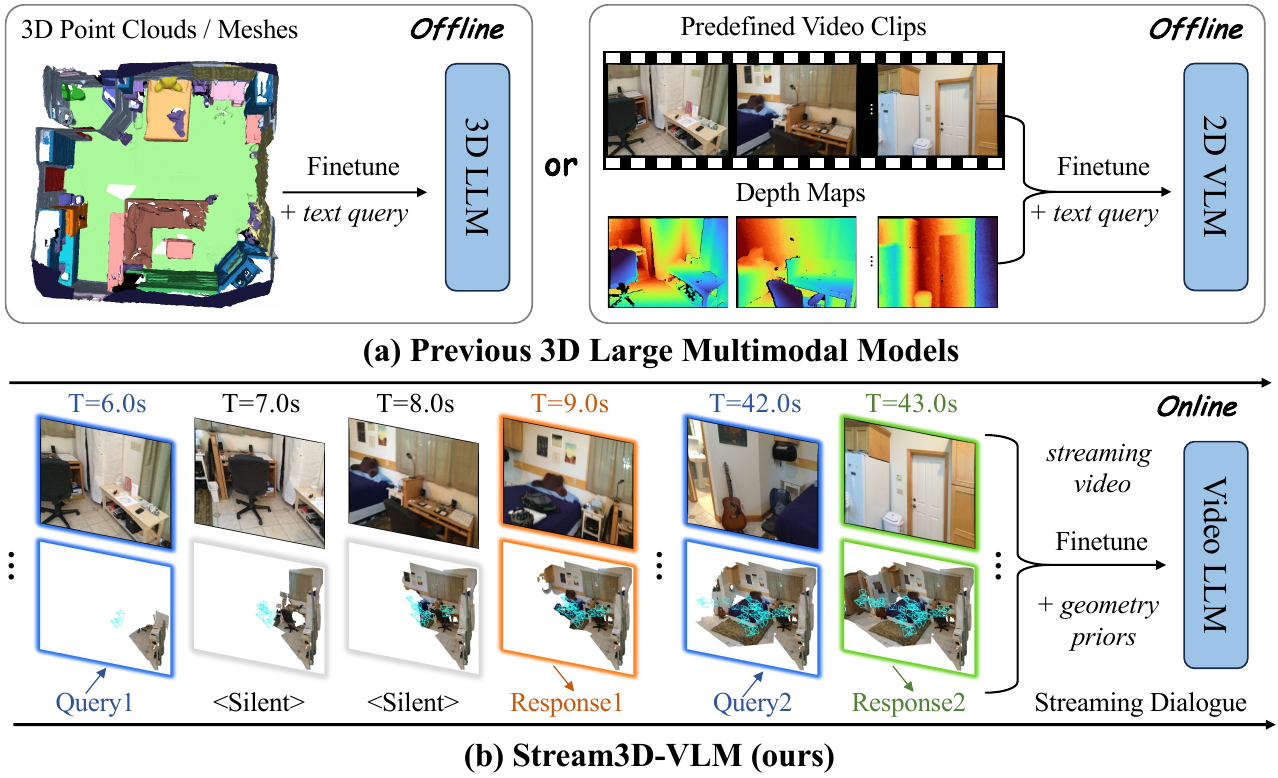}
  \caption{\textbf{Comparison between previous 3D LMMs and our Stream3D-VLM.} (a) Previous methods typically operate in offline settings, requiring complete 3D observations or predefined video clips. (b) In contrast, our method enables real-time 3D spatial understanding and interaction on streaming video by incrementally integrating geometry priors from StreamVGGT~\cite{zhuo2025streaming}.}
  \label{fig:teaser}
\end{figure}

With the rapid progress of Multimodal Large Language Models (MLLMs)~\cite{Qwen3-VL,wang2025internvl3.5,yu2026visiontrim,comanici2025gemini2.5,li2024llava-onevision,yu2026unlocking} in visual–language reasoning and interaction, growing attention has focused on extending them to more complex 3D vision tasks, such as autonomous robotics, AR/VR glasses, and embodied agents. The goal is to equip MLLMs with robust 3D spatial understanding, enabling them to function more effectively in real-world application scenarios.
Early pioneering works~\cite{deng20253d-llava,wang2025ross3d,kang2025robin3d,zhu2025llava-3d,huang20253drs} on 3D Large Multimodal Models rely on explicit 3D sensor inputs (\emph{e.g.}, point clouds, meshes, or depth maps) aligned with LLMs through instruction data. While effective at capturing geometric structure, their dependence on scarce 3D data prevents them from scaling to large-scale training, thereby limiting model capacity. Recent advances in feed-forward 3D reconstruction~\cite{lin2025depthv3,lan2025stream3r,yuan2026infinitevggt,su2026xstreamvggt,wang20254dvggt,wang2025vggt,wang2025cut3r} enable methods~\cite{zhao2025spacemind,fan2025vlm-3r} that require only RGB videos, injecting rich spatial information and scaling training on abundant 2D data. However, as shown in Figure~\ref{fig:teaser}, both paradigms remain strictly offline, requiring complete 3D scene observations or predefined video clips before interaction and task execution. In contrast, real-world embodied applications demand real-time interaction at arbitrary moments during video streaming.

Given prior efforts~\cite{chen2024videollm-online,tang2025time} in online 2D video understanding, which mainly target egocentric narration and forecasting, one might expect such models to extend to 3D tasks. However, 3D vision–language problems, including visual grounding and camera-related estimation, require deep reasoning over object-object and object-camera spatial relationships, \emph{i.e.}, the geometric structures inherent to the 3D world. Our experiments show that even with large-scale 3D–language fine-tuning, existing online 2D VLMs perform poorly on 3D tasks, highlighting the urgent need for a general-purpose 3D LMM capable of real-time interaction and powerful 3D spatial reasoning.

\begin{figure*}[t]
  \centering
  \includegraphics[width=\linewidth]{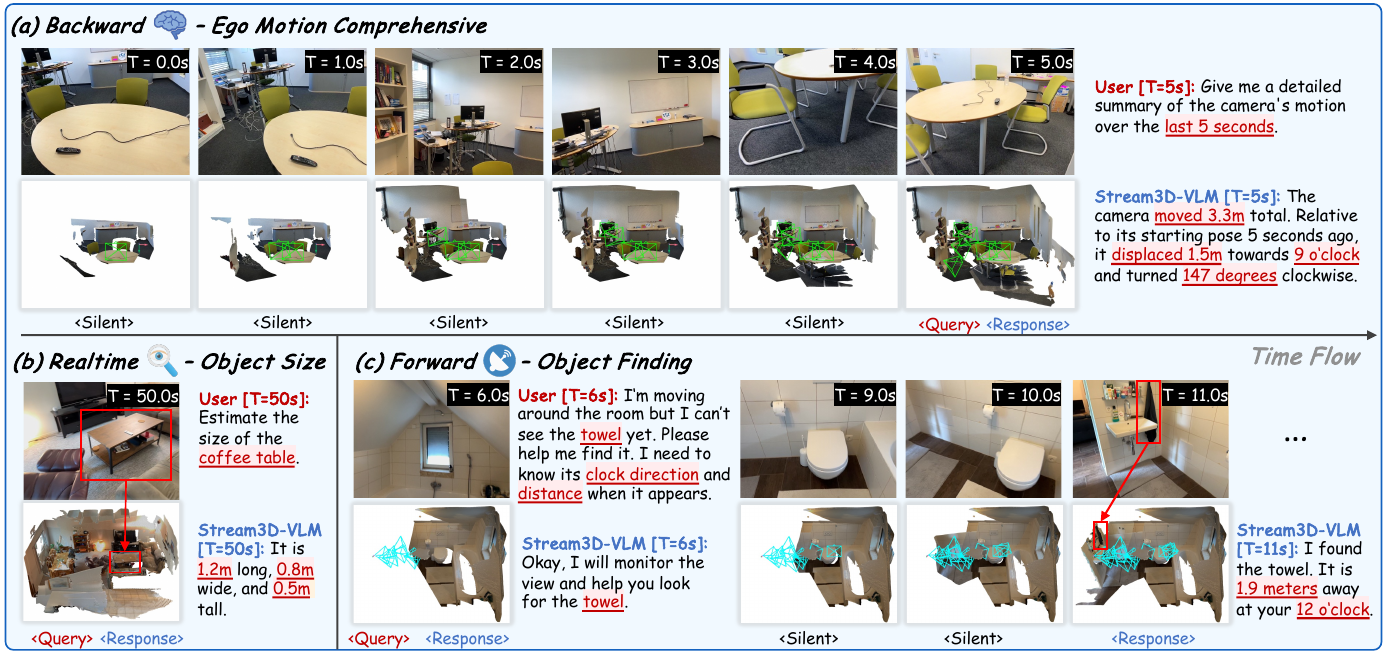}
\caption{\textbf{Qualitative examples of Stream3D-VLM on streaming videos from ScanNet++~\cite{yeshwanth2023scannet++}.} (a) \textbf{Backward:} The model aggregates historical visual-spatial features to precisely calculate the camera trajectory over a recent temporal window. (b) \textbf{Realtime:} The model performs metric reasoning to estimate the physical size of the coffee table in the current frame. (c) \textbf{Forward:} The model continuously monitors the video stream and responds proactively only when the target object becomes visible.}
  \label{fig:demo}
\end{figure*}

To address this challenge, we propose \textbf{Stream3D-VLM}, the first 3D vision–language model enabling online spatial understanding and interaction across multiple 3D–language tasks solely on streaming video. In this setting, the model is required to autonomously determine not only \textbf{\textit{what to answer}}, but also \textbf{\textit{when to respond}}, allowing natural and flexible interaction during continuous video input, as shown in Figure~\ref{fig:demo}. Built upon the LLM's native autoregressive training objective, we reformulate streaming control as a next-token prediction problem, enabling the model to learn response timing without degrading text generation quality. To support continuous 3D scene comprehension, we incrementally extract temporally aligned geometry priors by a streaming feed-forward 3D reconstruction model and inject them into the visual stream via a lightweight Visual–Spatial Feature Integration (VSFI) module. This design eliminates reliance on sparse 3D sensor data and supports scalable training on in-the-wild 2D videos. To handle long-term visual context during online inference, we introduce a plug-and-play Geometry-Adaptive Voxel Compression (GAVC) module that dynamically compresses visual tokens guided by spatial coordinates, preserving structural integrity while reducing redundancy and latency, enabling real-time deployment.

A key challenge in developing an online 3D spatial understanding model lies in the lack of large-scale streaming 3D–language data. Existing datasets for online 2D VLMs mainly target egocentric narration and activity recognition, offering limited supervision for 3D geometry and spatial relationships. To address this gap, we develop a scalable data generation pipeline to construct a large-scale online 3D spatio-temporal QA dataset with explicit timestamps, comprising over 1 million QA pairs across 5.2k videos. Furthermore, we introduce \textbf{Stream3D-Bench}, a novel benchmark consisting of 29 tasks over 518 videos, to systematically evaluate online 3D spatial understanding by measuring both model performance and temporal response accuracy.

Extensive experiments demonstrate that our approach achieves state-of-the-art online 3D spatial understanding and reasoning, while maintaining leading performance on offline tasks such as visual grounding and dense captioning. We believe this work lays a fundamental step toward deploying 3D LMMs in real-world embodied applications. To summarize, our main contributions are threefold:
\begin{itemize}
    \item We propose Stream3D-VLM, the first online 3D spatial understanding model solely on streaming video. By incrementally integrating geometry priors, it achieves effective 3D perception. A plug-and-play token compression module further reduces visual redundancy and inference latency, enabling real-time deployment.
    \item We develop a scalable data generation pipeline to curate online 3D spatio-temporal QA data with explicit timestamps for instruction tuning, addressing the scarcity of large-scale streaming 3D-language data.
    \item We establish Stream3D-Bench, a novel benchmark with 29 tasks across 518 videos to rigorously evaluate and advance online 3D spatial understanding.
\end{itemize}

\section{Related Work}
\textbf{3D Large Multimodal Models.} Build on advances in Large Language Models, recent works~\cite{hong20233d-llm,xu2024pointllm,fu2025scene-llm,hu2025g2,huang2025leo-vl,wang2025spatial3d,ouyang2025spacer,zhang2025spar,wang2025n3d} extend their capabilities to 3D data. Early methods~\cite{huang2024leo,yu2025inst3d,qi2025gpt4scene} extract features from point clouds, meshes, or depth maps and align them with LLMs via adapters. While effective at capturing geometric structure, they rely on sparse and costly 3D data, limiting the model's scalability and capacity. More recent methods~\cite{zheng2025vg-llm,wu2025spatial-mllm} rely solely on RGB videos and integrate spatial cues by geometry-aware encoders from 3D reconstruction models~\cite{wang2025vggt,wang2025cut3r}, eliminating reliance on explicit 3D sensors and enabling scalable training on 2D videos. However, both paradigms operate in offline settings, requiring complete 3D scenes or manually selected video clips, which diverges from real-world streaming scenarios. In this work, we propose an online 3D LMM framework that enables real-time interaction while maintaining strong 3D spatial understanding and reasoning, achieving leading performance across both online and offline 3D-language tasks.

\noindent\textbf{Online Video Understanding.} Driven by the growing real-time demands like AR glasses and autonomous robotics, recent research has advanced online video understanding, including action detection~\cite{yan2024action} and forecasting~\cite{wang2024anticipation}. VideoLLM-online~\cite{chen2024videollm-online} pioneers general-purpose LLMs for streaming video dialogue, while VideoLLM-MoD~\cite{wu2024videollm-mod} uses a mixture-of-depth strategy to scale resolution efficiently. StreamChat~\cite{liu2024streamchat} and VideoChat-Online~\cite{huang2025videochat-online} retain key video tokens via dynamic memory banks, and TimeChat-Online~\cite{yao2025timechat-online} and StreamingAssistant~\cite{jin2025streamingassistant} reduce temporal redundancy for faster decoding. Despite these advances, online 2D VLMs struggle with 3D tasks due to limited geometric and spatial understanding, emphasizing the need for a general-purpose online 3D LMM. By incrementally integrating geometry priors into the visual stream, Stream3D-VLM enables continuous and real-time 3D scene comprehension.

\section{Streaming 3D-Language Data Generation}
\label{sec:data_generation}

\begin{figure*}[t]
  \centering
  \includegraphics[width=\linewidth]{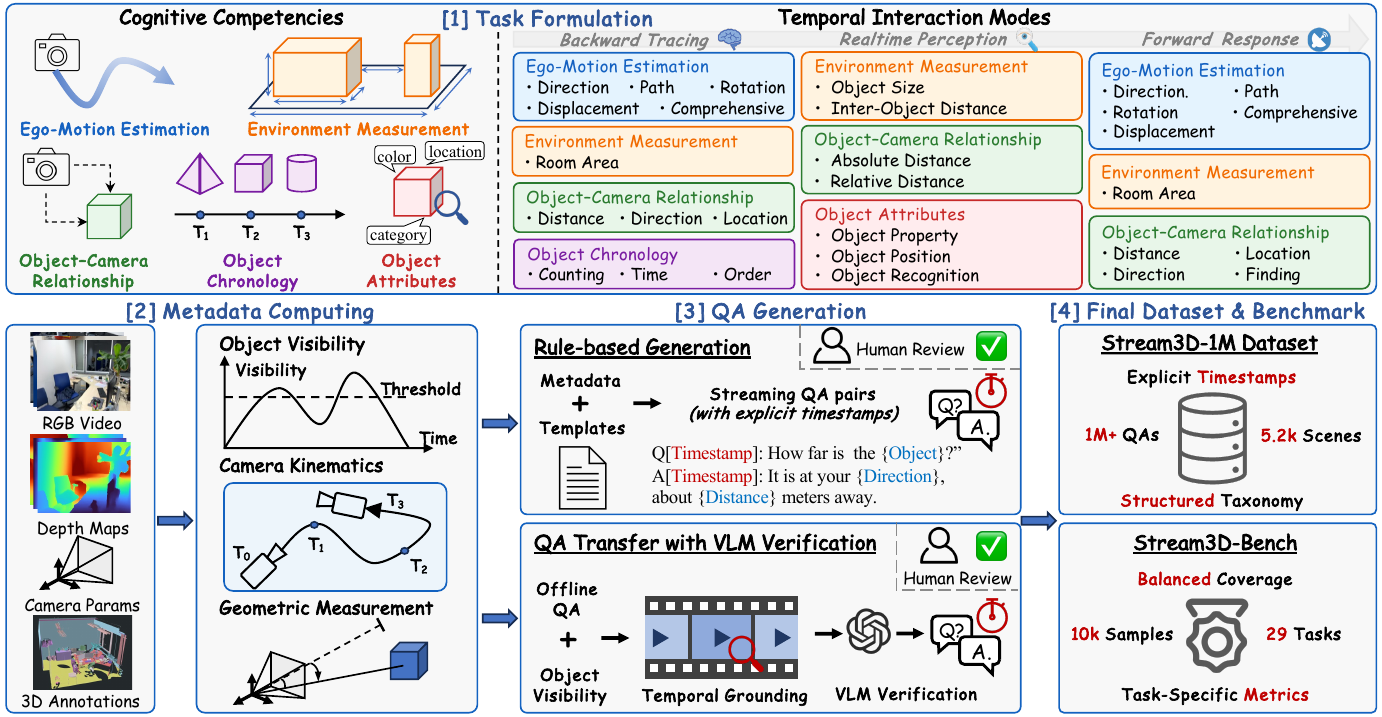}
\caption{\textbf{Illustration of our data generation pipeline.} Guided by a comprehensive task taxonomy spanning five cognitive competencies and three temporal interaction modes, the pipeline leverages detailed metadata from RGB-D video streams and a hybrid generation strategy to construct a large-scale spatio-temporal 3D QA dataset and the Stream3D-Bench for evaluating online 3D spatial understanding.}
  \label{fig:dataset_generation}
\end{figure*}

To endow our model to perceive, reason, and interact with 3D environments in an online setting, we introduce a scalable pipeline for generating large-scale spatio-temporal 3D-language data with explicit timestamps for instruction tuning. As shown in Figure~\ref{fig:dataset_generation}, we construct a comprehensive dataset of over 1 million QA pairs across 29 tasks spanning 5.2k 3D scans. Additionally, we carefully curate a high-quality benchmark, \textbf{Stream3D-Bench}, to assess the online 3D spatial understanding and reasoning of MLLMs.

\subsection{Task Taxonomy}
\label{subsec:taxonomy}
We define streaming 3D vision-language tasks along two orthogonal dimensions: \emph{Cognitive Competencies} and \emph{Temporal Interaction Modes}, forming a structured taxonomy that captures the core capabilities of an online 3D assistant.\\
\textbf{Cognitive Competencies.} 
We organize tasks along a hierarchical spectrum of spatial perception, ranging from ego-centric motion to global scene structure and fine-grained object-level analysis. \textit{Ego-Motion Estimation} captures the agent's own motion (\emph{e.g.}, path length and rotation angles). \textit{Environment Measurement} quantifies scene-level properties (\emph{e.g.}, room area and inter-object distances). \textit{Object–Camera Relationship} records spatial relations between the agent and objects, including distance and direction. \textit{Object Chronology} tracks objects over time, including counting, appearance order, and first/last observations. 
\textit{Object Attributes} describe object-level features, such as category, color, and location.\\
\textbf{Temporal Interaction Modes.} We define three interaction modes based on the relative query-answer timing. \textit{Backward Tracing (Memory)} retrieves information from past frames that is invisible in the current view, probing long-term memory capability. \textit{Realtime Perception (Observation)} grounds responses in visual evidence from the current frame, emphasizing immediate spatial perception. \textit{Forward Response (Monitoring)} entails asynchronous interaction, requiring the model to continuously monitor incoming frames and generate responses only when future conditions are satisfied.

\subsection{Data Collection and Processing}
\label{subsec:data_prep}

\textbf{Data Sources.} Our pipeline builds on the train splits of three widely used 3D datasets: ScanNet~\cite{dai2017scannet}, ScanNet++~\cite{yeshwanth2023scannet++}, and ARKitScenes~\cite{baruch2021arkitscenes}. Specifically, we utilize their RGB video streams, depth maps, camera parameters, and 3D instance segmentation annotations. Captured in real-world environments, these data preserve realistic characteristics such as motion blur, illumination changes, and sensor noise.\\
\textbf{Metadata Computing.} A key challenge in streaming 3D data generation lies in accurately grounding 2D frames in precise 3D properties. Hence, we develop a Core Annotation Engine that extracts structured metadata per frame. \textit{Object Visibility} is computed by projecting 3D geometry onto the image plane with depth-aware occlusion and temporally thresholding valid timestamps. In parallel, \textit{Camera Kinematics} is derived by analyzing the camera trajectory over time, including cumulative path length, net displacement, and horizontal rotation. \textit{Geometric Measurement} further captures spatial relationships like camera-object distances, relative azimuth, and global scene attributes. \textit{Further details on data generation are provided in the Supplementary Material.}

\subsection{Question-Answer Generation}
\label{subsec:generation_pipeline}
Building on the task taxonomy and computed metadata, we adopt a hybrid strategy combining geometric precision and semantic richness to curate 1M+ QA pairs across 29 tasks.\\
\textbf{Rule-based Generation.} For \textit{Ego-Motion Estimation}, \textit{Object–Camera Relationship}, \textit{Environment Measurement}, and \textit{Object Chronology}, we employ a template-based generator leveraging the computed metadata in Sec.~\ref{subsec:data_prep}. It processes time-series signals and fills natural-language templates to generate QA pairs. For example, in \textit{Object Chronology} tasks like Appearance Order, the generator extracts object entry timestamps and populates diverse linguistic templates.\\
\textbf{QA Transfer with VLM Verification.} For \textit{Object Attributes}, we transfer existing offline QAs~\cite{azuma2022scanqa} into the streaming setting via temporal grounding, linking each object-centric query to the frame $t^*$ where the object is sufficiently visible. We then apply VLM-based verification to ensure answerability under streaming conditions (\emph{e.g.}, motion blur or partial occlusion). Specifically, we use GPT-5~\cite{openai_gpt-5} to validate the visual evidence at $t^*$ and discard samples lacking adequate visual support.

\subsection{Stream3D-Bench}
\label{subsec:benchmark}

\textbf{Overview.} To evaluate MLLMs' online 3D spatial understanding from streaming video, we construct Stream3D-Bench using our QA generation pipeline. It comprises 10,037 manually curated high-quality samples spanning 518 real videos sourced from the validation sets of ScanNet, ScanNet++, and ARKitScenes. Stratified sampling ensures balanced coverage of all five content categories and three temporal interaction modes (Sec.~\ref{subsec:taxonomy}). As shown in Table~\ref{tab:compare_bench}, Stream3D-Bench significantly extends existing benchmarks such as VSI-Bench~\cite{yang2025vsi-bench} and OST-Bench~\cite{lin2025ost-bench} in both task diversity and question complexity. \textit{More visualizations are provided in the Supplementary Material.}

\noindent\textbf{Metrics.} Stream3D-Bench employs task-specific evaluation metrics. For Numerical Answers, we follow VSI-Bench and report \textit{Mean Relative Accuracy}. For Multiple-Choice Answers, we use \textit{Exact Match}. For Open-Ended Answers, we adopt an LLM-as-a-judge protocol with GPT-4o~\cite{hurst2024gpt-4o} to assess correctness. To evaluate the temporal precision inherent in streaming tasks, we introduce a new metric \textit{Answer-Timing Accuracy ($\mathcal{ATA}$)}, measuring how well the predicted response time aligns with the GT timestamp. Let $t_{\mathrm{pred}}$ denote the predicted response time and $t_{\mathrm{gt}}$ represent the earliest answerable GT timestamp. With a delay penalty factor $\beta=0.5$, the timing score is defined as
\begin{equation}
    S(t_{\mathrm{pred}}) = \mathbb{I}(t_{\mathrm{pred}} \ge t_{\mathrm{gt}}) \cdot \exp\bigl( -\beta (t_{\mathrm{pred}} - t_{\mathrm{gt}}) \bigr)
\end{equation}
$\mathcal{ATA}$ is computed by averaging the timing score over all samples. We additionally report Time-To-First-Token (TTFT), end-to-end latency, and memory usage to enable a comprehensive evaluation of inference efficiency. 

\begin{table}[t]
\centering
\caption{Comparison of our proposed Stream3D-Bench with existing benchmarks focusing on spatial intelligence in 3D worlds.}
\label{tab:compare_bench}
\resizebox{0.8\textwidth}{!}{%
\begin{tabular}{l|ccc}
\toprule
\textbf{Comparison} & \textbf{VSI-Bench} &  \textbf{OST-Bench} &  \textbf{Stream3D-Bench} \\
\midrule
\# Task Types & 8 & 15  & 29 \\
\# QA Pairs & 5k & 10k &  10k\\
Input Format  & Video Clips & Video Clips  & Streaming Video \\
Evaluation Granularity & Holistic & Holistic & Past / Present / Future \\
Response Timing Required & \xmark & \xmark  & \cmark \\
\bottomrule
\end{tabular}
}
\end{table}

\begin{figure*}[t]
  \centering
  \includegraphics[width=\linewidth]{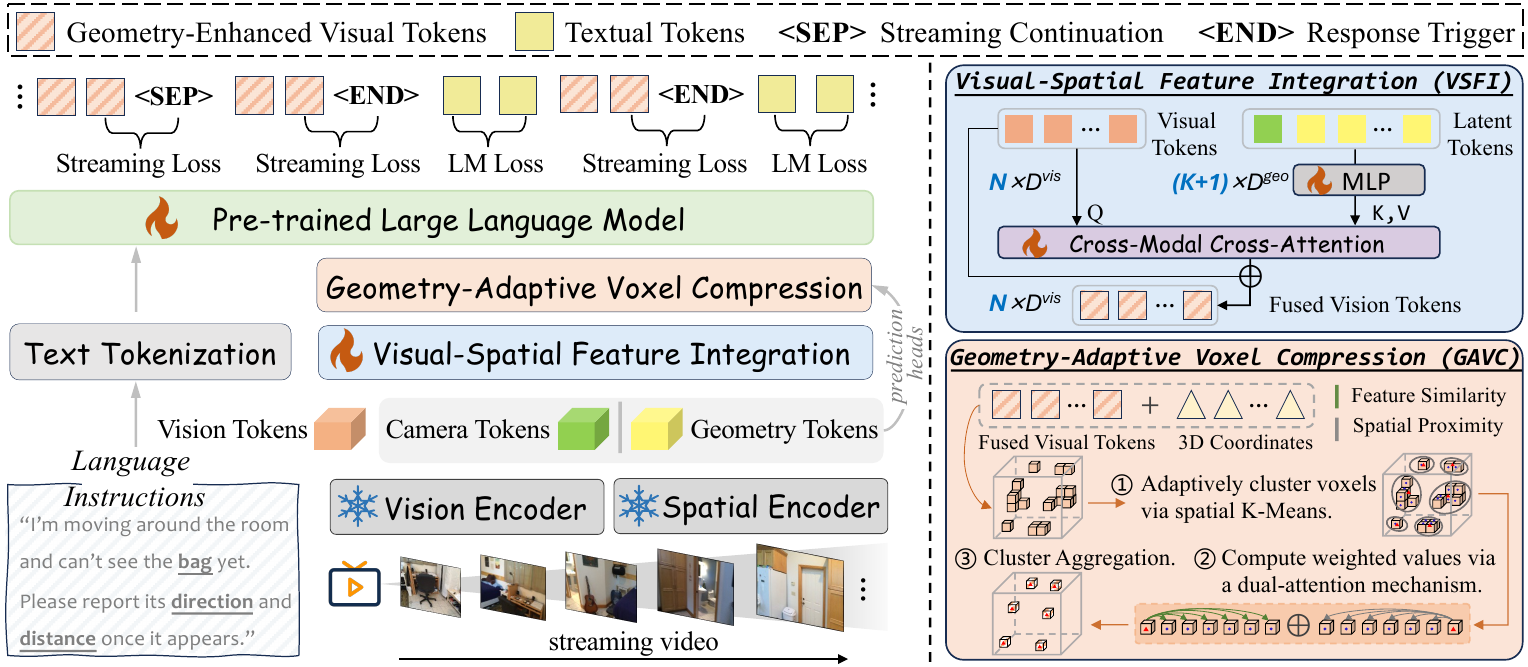}
  \caption{\textbf{Overview of our proposed Stream3D-VLM.} Our pipeline processes streaming video as a temporally ordered input sequence. We utilize the LLM’s native next-token prediction to jointly optimize a streaming control loss and the standard language modeling (LM) loss, enabling the model to learn when to respond or keep silent. We then suggest the VSFI module to inject temporally aligned geometric priors from a 3D reconstruction model into the visual stream. To mitigate long-context redundancy in online inference, we further propose a plug-and-play GAVC module that dynamically compresses visual tokens guided by 3D structure, enabling real-time deployment.}
  \label{fig:pipeline}
\end{figure*}

\section{Stream3D-VLM Architecture}
Our goal is to equip MLLM with online 3D spatial understanding and reasoning from streaming video, enabling diverse real-time visual interaction tasks. Figure~\ref{fig:pipeline} illustrates our framework. In this section, we first introduce a streaming decision learning strategy that utilizes the LLM's native next-token prediction objective to model response timing after a user query. Next, we present the Visual-Spatial Feature Integration (\textbf{VSFI}) module to inject geometry priors into the visual stream for continuous 3D scene comprehension. Finally, we propose a plug-and-play Geometry-Adaptive Voxel Compression (\textbf{GAVC}) module that dynamically compresses visual tokens guided by spatial coordinates, preserving structural integrity while significantly reducing redundancy and latency during long-context online inference. 

\subsection{Autoregressive Streaming Mechanism}
Existing 3D LMMs are primarily designed for offline inference on pre-segmented video clips. When deployed in streaming scenarios, they necessitate repeatedly reprocessing the entire historical context at each incoming frame to decide whether to respond, leading to excessive memory usage and high inference latency. This inefficiency severely limits their applicability in real-time settings.

\noindent\textbf{Streaming Control Modeling.} Inspired by recent advances in online 2D video LLMs~\cite{chen2024videollm-online,wu2024videollm-mod}, we reformulate streaming control as a next-token prediction problem. Specifically, we exploit the LLM's native autoregressive objective to enable the model to learn \textit{when to answer}—skipping redundant frames and triggering responses only when necessary, rather than generating outputs at every time step. To this end, we introduce two special decision tokens, \textbf{\texttt{<SEP>}} and \textbf{\texttt{<END>}}, which indicate whether the model should continue ingesting visual inputs or stop and initiate response generation, as illustrated below:
\begin{equation}\label{equ:streaming_illustration}
\resizebox{0.57\linewidth}{!}{$
\begin{split}
    &\underbrace{\texttt{USER:}\,\texttt{<img>}\,\texttt{Query}}_{\text{Context History}} \underbrace{\texttt{<img>}\,\textbf{\texttt{<SEP>}}\,\texttt{<img>}\,\textbf{\texttt{<SEP>}}}_{\text{Streaming Continuation}} \\ 
    &\underbrace{\mathstrut \texttt{<img>}\,\textbf{\texttt{<END>}}}_{\text{Response Trigger}} \
\underbrace{\mathstrut \texttt{ASSISTANT: <txt>\,...\,<txt>}}_{\text{Response Generation}}
\end{split}
$}
\end{equation}

\noindent\textbf{Joint Training Objective.} To unify streaming control and language generation within a single autoregressive framework, we modify the training label mask to supervise these decision tokens explicitly. The overall training objective is a weighted sum of the streaming decision loss $\mathcal{L}_{\text{stream}}$ and the standard language modeling loss $\mathcal{L}_{\text{LM}}$:
\begin{equation}
\begin{gathered}
\mathcal{L}_{\text{stream}} = \frac{1}{|\mathcal{D}|} \sum_{t \in \mathcal{D}} \mathrm{CE}_t, \quad
\mathcal{L}_{\text{LM}} = \frac{1}{|\mathcal{T}|} \sum_{t \in \mathcal{T}} \mathrm{CE}_t, \\
\mathrm{CE}_t \equiv \mathrm{CE}\bigl(p_\theta(y_t \mid y_{<t}), y_t\bigr), \
\mathcal{L} = \lambda\,\mathcal{L}_{\text{stream}} + \mathcal{L}_{\text{LM}},
\end{gathered}
\end{equation}
where $\mathcal{D}$ is the set of streaming decision tokens $y_t \in \{\textbf{\texttt{<SEP>}}, \textbf{\texttt{<END>}}\}$, $\mathcal{T}$ contains all remaining tokens, $\mathrm{CE}(\cdot)$ denotes the cross-entropy loss, $p_\theta$ is the model’s next-token prediction distribution, and $\lambda$ balances the two objectives. During inference, the model decides whether to respond at each frame via next token prediction (\textbf{\texttt{<SEP>}} or \textbf{\texttt{<END>}}).

\subsection{Visual-Spatial Feature Integration}

To endow MLLMs with spatial understanding without explicit 3D inputs, we use latent geometry from feed-forward 3D reconstruction models, enabling scalable training on large-scale 2D videos. Specifically, we adopt StreamVGGT~\cite{zhuo2025streaming} to incrementally extract geometry priors from streaming video in online settings.

\noindent\textbf{Latent Geometric Encoding.} Given an incoming RGB frame $I_t \in \mathbb{R}^{3 \times H \times W}$, we first extract 2D visual tokens $\mathbf{H}^{\mathrm{2D}}_t \in \mathbb{R}^{N \times D^{\textit{vis}}}$ via the MLLM's native vision encoder. In parallel, StreamVGGT's spatial encoder produces latent geometry tokens $\mathbf{G}_t \in \mathbb{R}^{K \times D^\textit{geo}}$ along with a camera token $\mathbf{c}_t \in \mathbb{R}^{1 \times D^\textit{geo}}$ encoding global camera and scene information. We concatenate these tokens and project them into the LLM embedding space via a lightweight two-layer MLP:
\begin{equation}
\mathbf{H}_t^{\mathrm{3D}}
=
\mathrm{MLP}\!\left(
\left[
\mathbf{c}_t \,;\, \mathbf{G}_t
\right]
\right)
\in \mathbb{R}^{(K+1) \times D^\textit{vis}}.
\label{eq:geometry_projection}
\end{equation}

\noindent\textbf{Cross-Attention Fusion.} To inject geometric priors into the semantic stream, we treat
$\mathbf{H}^{\mathrm{2D}}_t$
as queries and
$\mathbf{H}_t^{\mathrm{3D}}$
as keys and values.
A stack of cross-attention blocks yields geometry-enhanced visual tokens
$\mathbf{H}^{\mathrm{f}}_t \in \mathbb{R}^{N \times D^{\textit{vis}}}$:
\begin{equation}
\mathbf{H}^{\mathrm{f}}_t
=
\mathrm{softmax}\!\left(
\frac{
(W_Q\!\mathbf{H}^{\mathrm{2D}}_t)
(W_K\!\mathbf{H}^{\mathrm{3D}}_t)^{\top}
}{
\sqrt{d_k}
}
\right)\!
(W_V\!\mathbf{H}^{\mathrm{3D}}_t),
\end{equation}
where $W_Q$, $W_K$, and $W_V$ are learnable projections and $d_k$ denotes the key dimension. A residual connection is applied to preserve the original semantics as $\mathbf{H}^{\mathrm{f}}_t \leftarrow \mathbf{H}^{\mathrm{f}}_t + \mathbf{H}^{\mathrm{2D}}_t$.

\subsection{Geometry-Adaptive Voxel Compression}
After visual-spatial feature integration, we introduce a plug-and-play module that dynamically compresses visual tokens guided by spatial coordinates, effectively reducing long-context visual redundancy during online inference.

\noindent\textbf{3D Voxel Construction.} For each incoming frame $I_t \in \mathbb{R}^{3 \times H \times W}$, we first use StreamVGGT's prediction heads to estimate the depth map $D_t \in \mathbb{R}^{H \times W}$ and camera intrinsics and extrinsics $(\mathbf{K}_t,\mathbf{E}_t )$. Given the 2D patch coordinates $(u_j, v_j)$ in $I_t$, each patch is back-projected to 3D position:
\begin{equation}
    \mathbf{p}_{t,j} = \mathbf{E}_t^{-1} \left( D_t(u_j, v_j) \mathbf{K}_t^{-1} [u_j, v_j, 1]^\top \right).
    \label{eq:back_projection}
\end{equation}
The 2D tokens are then lifted into spatially-aware 3D voxels via sinusoidal positional encoding: $\mathbf{v}_{t,j} = \mathbf{H}^{\mathrm{f}}_{t, j} + \mathrm{PE}(\mathbf{p}_{t,j})$.

\noindent\textbf{Dynamic Clustering.} To enable dynamic compression while preserving the inherent 3D structure, the newly constructed voxels at time $t$ are collected as:
\begin{equation}
    \mathcal{V}_t
=
\{(\mathbf{v}_{t,j}, \mathbf{p}_{t,j})\}_{j=1}^{N}.
\end{equation}
We then apply spatial K-Means clustering to adaptively partition the combined voxel set into $K$ clusters:
\begin{equation}
\{\mathcal{C}_k\}_{k=1}^{K}
=
\mathrm{KMeans}
\!\left(
\{\mathbf{p} \mid (\mathbf{v}, \mathbf{p}) \in \mathcal{V}_t\},
\, K
\right),
\label{eq:spatial_kmeans}
\end{equation}
where each cluster $\mathcal{C}_k$ groups spatially proximal voxels in 3D space, enabling structure-aware dynamic compression. The clustering process is executed in parallel on the GPU, incurring negligible additional latency.

\noindent\textbf{Dual-Attention Aggregation.} Within each cluster $\mathcal{C}_k$, voxel features are aggregated via a dual-attention mechanism that models both \emph{feature similarity} and \emph{spatial proximity}. The cluster center's feature and coordinate are computed as:
\begin{equation}
\bar{\mathbf{v}}_k
=
\frac{1}{|\mathcal{C}_k|}
\sum_{j \in \mathcal{C}_k}
\mathbf{v}_j , \ \bar{\mathbf{p}}_k
=
\frac{1}{|\mathcal{C}_k|}
\sum_{j \in \mathcal{C}_k}
\mathbf{p}_j .
\end{equation}
Each voxel $\mathbf{v}_j \in \mathcal{C}_k$ is assigned feature and spatial weights:
\begin{align}
s_j^{\mathrm{f}} = \cos(\mathbf{v}_j, \bar{\mathbf{v}}_k), \
s_j^{\mathrm{p}} =
\exp\!\left(
-\|\mathbf{p}_j - \bar{\mathbf{p}}_k\|^2 / {2\sigma_k^2}
\right),
\end{align}
where $\sigma_k$ is the voxel distance standard deviation to the center. The two are combined as $w_j
=
\alpha\, s_j^{\mathrm{f}}
+
(1-\alpha)\, s_j^{\mathrm{p}}.
$

The aggregated feature for cluster $k$ is $\mathbf{v}'_k = \sum_{j \in \mathcal{C}_k} w_j \mathbf{v}_j$. These compressed voxels $\{\mathbf{v}'_k\}_{k=1}^{K}$, encoding both rich semantics and spatial cues, are then fed to the LLM with prompt tokens for efficient autoregressive generation.

\section{Experiments}
\subsection{Experimental Setup}
\textbf{Datasets and Benchmarks.} To evaluate our model across online and offline 3D-language tasks, we train it on the curated 1M+ streaming 3D spatio-temporal QA pairs, alongside the VSI-590K dataset~\cite{yang2025cambrian-s} including real videos from the train splits of S3DIS, ScanNet, ScanNet++, ARKitScenes, and Aria Digital Twin. For evaluation, we employ Stream3D-Bench for online spatial understanding, VSI-Bench~\cite{yang2025vsi-bench} for offline spatial reasoning, and downstream 3D scene understanding tasks, including ScanRefer~\cite{chen2020scanrefer} for visual grounding, ScanQA~\cite{azuma2022scanqa} for question answering, and Scan2Cap~\cite{chen2021scan2cap} for dense captioning.\\
\textbf{Implementation Details.} Our model is based on Qwen2.5-VL-3B/7B~\cite{Qwen2.5-VL} and integrates StreamVGGT-1B~\cite{zhuo2025streaming} as the spatial encoder to supply incremental 3D geometry priors temporally aligned with the streaming video input. We train our model in a unified end-to-end multi-task instruction tuning paradigm for a single epoch on the mixed dataset. We adopt the AdamW optimizer with a weight decay of 0.03, using randomly sampled data at each step. During training, the MLLM visual encoder and spatial encoder are frozen, while the proposed VSFI module and LLM backbone are fully trainable.\\ 
\textbf{Inference Pipeline.} During online inference, the video is input as a frame-by-frame stream with a default 1 FPS. Our model ingests each frame sequentially and generates tokens on-the-fly. In the whole process, KV caching is employed to accelerate decoding, implicitly reusing previously generated tokens without explicit concatenation across frames.

\begin{table*}[t]
    \caption{\textbf{Evaluation results on Stream3D-Bench.} Stream3D-VLM consistently outperforms all competing models, delivering the most accurate response timing and the lowest inference latency. NA/MCA/OEA denote numerical, multiple-choice, and open-ended answers, respectively. FT indicates that the model is fine-tuned on our curated online 3D spatio-temporal QA dataset for fair comparison. \textbf{Bold} and \underline{underlined} values indicate the best and the second-best results, respectively. The results are reported under a 1fps streaming video setting.}
    \label{tab:stream3d-bench}
    \vspace{-1mm}
    \setlength{\tabcolsep}{1pt}
    \centering
    \renewcommand\arraystretch{1.2}
    \resizebox{\textwidth}{!}{
    \begin{tabular}{r|c|ccccccccc|crcc|c}
        \toprule
        \multirow{2}{*}{\textbf{Methods}} & \multirow{2}{*}{{Avg.} \textuparrow}& \multicolumn{3}{c}{\cellcolor{yellow!10}{\textbf{Backward Tracing}}} & \multicolumn{3}{c}{\cellcolor{green!10}{\textbf{Realtime Perception}}} & \multicolumn{3}{c|}{\cellcolor{orange!10}{\textbf{Forward Response}}} & \multirow{2}{*}{\makecell[c]{Answer- \\ Timing Acc.}\textuparrow} & \multirow{2}{*}{\makecell[c]{TTFT\\Latency}\textdownarrow} & \multirow{2}{*}{\makecell[c]{End2End\\Latency}\textdownarrow} & \multirow{2}{*}{\makecell[c]{Memory\\Usage}\textdownarrow} & \multirow{2}{*}{\makecell[c]{Image\\Resolution}}\\
        \cmidrule(lr){3-5} \cmidrule(lr){6-8} \cmidrule(lr){9-11}
        &  & NA & MCA & OEA & NA & MCA & OEA & NA & MCA & OEA & & & & &  \\
        \hline
        \rowcolor{navyblue!10}
        \multicolumn{1}{l|}{\textit{{Proprietary Models (API)}}} & & & & & & & & & & & & & & & \\
        GPT-4o  & 28.0 & 10.5 & 36.3 & 24.9 & 31.1 & 34.8 & 51.1 & 10.2 & 29.9 & 23.0 & 55.9\% & -- & -- & -- & 1296$\times$968\\
        GPT-5 & 35.0 & 18.7 & 44.1 & 33.6 & 38.0 & 46.2 & \underline{52.8} & 12.9 & 30.6 & 37.7 & 61.7\% & -- & -- & -- & 1296$\times$968 \\
        \hline
        \rowcolor{navyblue!10}
        \multicolumn{1}{l|}{\textit{{Open-source Models}}} & & & & & & & & & & & & & & & \\
        LLaVA-Video-7B  & 20.4 & 12.7 & 34.6 & 11.4 & 21.6 & 28.6 & 38.0 & 10.2 & 20.3 & 6.5 & 55.5\% & 650ms & 2.84s & \underline{25.8G} & 1152$\times$768\\
        InternVL3-8B & 24.1 & 22.1 & 34.4 & 6.8 & 30.7 & 31.2 & 40.0 & 18.3 & 27.2 & 5.9 & 19.8\% & 1034ms & 5.65s & 40.5G & 1024$\times$765\\
        InternVL3.5-8B   & 27.0 & 21.4 & 36.2 & 17.6 & 28.2 & 38.2 & 42.4 & 25.2 & 25.8 & 8.2 & 23.0\% & 1771ms & 6.72s & 43.1G & 1024$\times$765\\
        Qwen2.5-VL-7B   & 22.5 & 11.6 & 36.3 & 13.6 & 23.1 & 33.2 & 39.4 & 11.3 & 28.4 & 5.5 & 19.2\% & 325ms & 3.16s & 40.0G & 1008$\times$784 \\
        Qwen2.5-VL-32B  & 29.3 & 14.1 & 40.3 & 35.7 & 28.7 & 37.6 & 43.2 & 22.4 & 30.2 & 11.5 & 35.2\% & 560ms & 5.87s & 91.6G & 1008$\times$784\\
        Qwen3-VL-8B   & 19.0 & 2.5 & 33.2 & 20.2 & 16.0 & 33.4 & 28.4 & 4.6 & 24.7 & 8.3 & 61.7\% & 150ms & 1.56s & 28.8G & 1024$\times$768\\
        Qwen3-VL-32B  & 23.1 & 4.9 & 34.2 & 35.9 & 19.4 & 32.0 & 31.8 & 7.3 & 29.2 & 13.5 & 66.4\% & 282ms & 3.22s & 65.5G & 1024$\times$768\\

        \hline
        \rowcolor{navyblue!10}
        \multicolumn{1}{l|}{\textit{{Streaming Dialogue VLMs}}} & & & & & & & & & & & & & & & \\
        VideoLLM-online-8B (FT) & 34.6 & 28.8 & 40.9 & 27.8 & 31.3 & 37.9 & 44.6 & 38.2 & 32.7 & 29.2 & 70.2\% & 120ms & 1.29s & 28.4G & 384$\times$384\\
        Qwen2.5-VL-7B (FT) & 47.8 & 45.6 & 56.6 & \underline{45.4} & 45.7 & 44.6 & 48.7 & 46.6 & 57.4 & 39.8 & 73.1\% & 106ms & 0.53s & 36.8G & 504$\times$392\\
        \textbf{Stream3D-VLM-4B}  & \underline{54.6} & \underline{58.1} & \underline{61.2} & 45.2 & \underline{58.8} & \underline{57.3} & \underline{52.8} & \underline{55.3} & \underline{60.6} & \underline{41.9} & \underline{75.4\%} & \textbf{43ms} & \textbf{0.24s} & \textbf{20.7G} &  504$\times$392\\
        \textbf{Stream3D-VLM-8B} & \textbf{58.8} & \textbf{59.8} & \textbf{67.9} & \textbf{50.5} & \textbf{61.4} & \textbf{61.5} & \textbf{54.3} & \textbf{58.1} & \textbf{66.7} & \textbf{49.4} & \textbf{86.7\%} & \underline{62ms} & \underline{0.39s} & 36.6G & 504$\times$392\\
        \hline
    \end{tabular}
    }
    
\end{table*}

\begin{table*}[t]
    \caption{\textbf{Evaluation results on VSI-Bench.} Despite being designed for streaming scenarios, Stream3D-VLM also performs well across all subtasks of the offline spatial perception and reasoning benchmark, significantly surpassing both commercial and open-source models.}
    \label{tab:vsibench}
    \vspace{-1mm}
    \centering
    \renewcommand\arraystretch{0.9}
    \resizebox{\textwidth}{!}{
    \begin{tabular}{rc|c|cccccccc}
        \toprule
        \multirow{2}{*}{\textbf{Methods}} & \multirow{2}{*}{\textbf{Online}} & \multirow{2}{*}{{Avg.} \textuparrow} & \multicolumn{4}{c}{\textbf{Numerical Answer}} & \multicolumn{4}{c}{\textbf{Multiple-Choice Answer}} \\
        \cmidrule(lr){4-7} \cmidrule(lr){8-11}
        &  &  & Obj. Cnt. & Abs. Dist. & Obj. Size & Room Size & Rel. Dist. & Rel. Dir. & Route Plan & Appr. Order\\
        \hline
        \rowcolor{navyblue!10}
        \multicolumn{1}{l}{\textit{{Proprietary Models (API)}}} & & & & & & & & & & \\
        GPT-4o & \xmark  & 34.0 & 46.2 & 5.3 & 43.8 & 38.2 & 37.0 & 41.3 & 31.5 & 28.5 \\
        Gemini-1.5 Pro & \xmark  & 45.4 & 56.2 & 30.9 & 64.1 & 43.6 & 51.3 & 46.3 & 36.0 & 34.6 \\
        Gemini-2.5 Pro & \xmark  & 51.5 & 43.8 & 34.9 & 64.3 & 42.8 & 61.1 & 47.8 & \underline{45.9} & \underline{71.3} \\
        \hline
        \rowcolor{navyblue!10}
        \multicolumn{1}{l}{\textit{{Open-source Models}}} & & & & & & & & & & \\
        LongVILA-8B & \xmark & 21.6 & 29.1 & 9.1 & 16.7 & 0.0 & 29.6 & 30.7 & 32.5 & 25.5\\
        VILA-1.5-40B & \xmark & 31.2 & 22.4 & 24.8 & 48.7 & 22.7 & 40.5 & 25.7 & 31.5 & 32.9\\
        Qwen2.5-VL-7B & \xmark & 33.0 & 40.9 & 14.8 & 43.4 & 10.7 & 38.6 & 38.5 & 33.0 & 29.8\\
        Qwen2.5-VL-72B & \xmark & 37.0 & 25.1 & 29.3 & 54.5 & 38.8 & 38.2 & 37.0 & 34.0 & 28.9 \\
        LLaVA-OneVision-72B  & \xmark  & 40.2 & 43.5 & 23.9 & 57.6 & 37.5 & 42.5 & 39.9 & 32.5 & 44.6\\
        LLaVA-NeXT-Video-72B & \xmark  & 40.9 & 48.9 & 22.8 & 57.4 & 35.3 & 42.4 & 36.7 & 35.0 & 48.6 \\
        \hline
        \rowcolor{navyblue!10}
        \multicolumn{1}{l}{\textit{{Spatial Reasoning Models}}}& & & & & & & & & & \\
        SpaceR-7B & \xmark & 45.5 & 57.8 & 28.2 & 59.9 & 47.1 & 40.1 & 45.4 & 33.5 & 52.1 \\
        Spatial-MLLM-4B & \xmark & 48.4 &  65.3 & 34.8 & 63.1 & 45.1 & 41.3 & 46.2 & 33.5 & 46.3\\
        VG LLM-8B  & \xmark & 50.7 & 67.9 & 37.7 & 58.6 & 62.0 & 46.6 & 40.7 & 32.4 & 59.2\\
        VLM-3R-8B & \xmark & \underline{60.9} & \underline{70.2} & \underline{49.4} & \underline{69.2} & \underline{67.1} & \underline{65.4} & \textbf{80.5} & 45.4 & 40.1 \\
        \textbf{Stream3D-VLM-4B} & \cmark & 55.2 & 68.6 & 38.4 & 65.9 & 56.4 & 49.9 & 54.3 & 42.3 & 66.0 \\
        
        \textbf{Stream3D-VLM-8B} & \cmark & \textbf{65.9} & \textbf{72.4} & \textbf{50.1} & \textbf{71.5} & \textbf{68.7} & \textbf{70.8} & \underline{73.3} & \textbf{46.2} & \textbf{73.8} \\
        \hline
    \end{tabular}
    }
\end{table*}

\begin{table*}[t]
    \caption{\textbf{Evaluation results on ScanQA, ScanRefer, and Scan2Cap.} Stream3D-VLM, while operating in the online setting, excels in traditional 3D scene understanding tasks, including Question Answering, Visual Grounding, and Dense Captioning.}
    \label{tab:scanqa+scanrefer+scan2cap}
    \vspace{-1mm}
    \centering
    \renewcommand\arraystretch{0.95}
    \resizebox{\textwidth}{!}{
    \begin{tabular}{rc|ccccccccccc}
        \toprule
        \multirow{2}{*}{\textbf{Methods}} & \multirow{2}{*}{\textbf{Online}} & \multicolumn{5}{c}{\textbf{ScanQA}} & \multicolumn{2}{c}{\textbf{ScanRefer}} & \multicolumn{4}{c}{\textbf{Scan2Cap@0.50}} \\
        \cmidrule(lr){3-7} \cmidrule(lr){8-9} \cmidrule(lr){10-13}
        &  & BLEU-4  & METEOR  & ROUGE & CIDEr  & EM  & Acc@0.25  & Acc@0.50  & BLEU-4  & METEOR & ROUGE  & CIDEr \\
        \hline
        \rowcolor{navyblue!10}
        \multicolumn{1}{l}{\textit{{Task-Specific Models}}} & & & & & & & & & & & & \\
        ScanQA   & \xmark & 10.1 & 13.1 & 33.3 & 64.9 & 21.0 & -- & -- & -- & -- & -- & -- \\
        ScanRefer & \xmark & -- & -- & -- & -- & -- & 37.3 & 24.3 & -- & -- & -- & -- \\
        Scan2Cap & \xmark & -- & -- & -- & -- & -- & -- & -- & 22.4 & 21.4 & 43.5 & 35.2 \\
        3D-Vista  & \xmark & 13.1 & 15.2 & 38.6 & 76.6 & 27.0 & 50.6 & 45.8 & 34.0 & 27.1 & 54.3 & 66.9 \\
        \hline
        \rowcolor{navyblue!10}
        \multicolumn{1}{l}{\textit{{3D/2.5D-Input Models}}} & & & & & & & & &  & & & \\
        3D-LLM  & \xmark & 12.0 &14.5 & 35.7& 69.4& 20.5& 30.3 & -- &8.1& 13.1&33.2 & --\\
        LEO  & \xmark & 11.5 & 16.2 & 39.3 & 80.0 &24.5 & -- & -- & 38.2 & 27.9 & 58.1 & 72.4 \\
        Inst3D-LMM & \xmark & 14.9 & 18.4 & 42.6 & 88.6 & 24.6 & 57.8 & 51.6 & 38.3 & 27.5 & 57.2 & 79.7\\
        LLaVA-3D & \xmark & 14.5 & \underline{20.7} & \underline{49.6} & 91.7 & 27.0 &54.1 & 42.4 & 41.1 & \underline{30.2} & 63.4 & 79.2\\
        Video-3D LLM & \xmark  & 16.2 & 19.8 & 49.0 & 102.1 & \underline{30.1} & \underline{58.1} & 51.7 & 40.2 & 28.5 & 61.7 & \underline{80.0} \\
        \hline
        \rowcolor{navyblue!10}
        \multicolumn{1}{l}{\textit{{Only Video-Input Models}}} & & & & & & & & &  & & & \\
        SPAR-8B & \xmark & 15.3 & -- & -- & 90.7 & 27.7 & 48.8 & 43.1 & -- & -- & -- & -- \\
        Spatial-MLLM-4B & \xmark & 14.8 & 18.4 & 45.0 & 91.8 & 26.3 & 49.3 & 44.2 & 37.8 & 27.3 & 57.8 & 76.5\\
        VG LLM-8B & \xmark & 16.0 & 18.0 & 44.2 & 98.6 & 27.3& 54.4 & 47.9 & 40.1 & 28.4& 62.0 & 76.4\\
        \textbf{Stream3D-VLM-4B} & \cmark & \underline{16.6} & 19.5 & 47.9 & \underline{102.6} & 29.2 & 56.5 & \underline{51.9} & \underline{41.6} & 28.8 & \underline{63.5} & 78.8\\
        
        \textbf{Stream3D-VLM-8B}  & \cmark & \textbf{17.8} & \textbf{21.0} & \textbf{50.2} & \textbf{104.5} & \textbf{30.9} & \textbf{58.4} & \textbf{52.5} & \textbf{42.8} & \textbf{31.0} & \textbf{64.2} & \textbf{81.2} \\
        \hline
    \end{tabular}
    }
    \vspace{-2mm}
\end{table*}

\subsection{Main Results}

\textbf{Evaluation on Stream3D-Bench.} Since existing offline models cannot natively process streaming videos, we provide the full video and specify the query time, prompting joint prediction of response timing and answers. We also fine-tune an online 2D VLM~\cite{chen2024videollm-online} and Qwen2.5-VL-7B with streaming decision learning on the curated 3D QA pairs as baselines. As shown in Table~\ref{tab:stream3d-bench}, our method achieves strong spatial understanding and temporal reasoning across three categories, significantly outperforming both proprietary and open-source models. 

\noindent\textbf{Evaluation on VSI-Bench.} To examine whether streaming prediction affects offline performance, we evaluate our method on VSI-Bench. As shown in Table~\ref{tab:vsibench}, our 8B model achieves the highest average accuracy of 65.9\%, outperforming both open-source general models and specialized spatial reasoning models. Notably, the 4B variant reaches 55.2\% accuracy, surpassing much larger 72B models and the proprietary model Gemini-2.5 Pro~\cite{comanici2025gemini2.5}. These results demonstrate our method’s strong 3D spatial understanding and reasoning, even in offline settings.

\noindent\textbf{Evaluation on ScanQA, ScanRefer, and Scan2Cap.} We also evaluate Stream3D-VLM on other 3D scene understanding tasks—question answering, visual grounding, and dense captioning in Table~\ref{tab:scanqa+scanrefer+scan2cap}. Despite using online videos without explicit 3D data, our method outperforms others across all tasks and metrics. The 4B variant remains competitive, exceeding many larger methods.

\begin{table}[t]
\centering
\caption{\textbf{Ablation of streaming loss design.} We explore different loss functions and weighting factors to identify an optimal trade-off across multiple evaluation metrics under the online input setting.}
\label{tab:ablation_streamingloss}
\setlength{\tabcolsep}{8pt}
\renewcommand\arraystretch{0.99}
\resizebox{0.8\textwidth}{!}{%
\begin{tabular}{c|ccc|c|c}
\toprule
\multirow{2}{*}{\textbf{Configs}} & \multicolumn{3}{c|}{\textbf{Stream3D-Bench}} & \multirow{2}{*}{\textbf{\makecell[c]{Answer- \\ Timing Acc.}\textuparrow}} &  \multirow{2}{*}{\textbf{\makecell[c]{End2End \\ Latency} \textdownarrow}} \\
\cmidrule(lr){2-4}
& NA \textuparrow & MCA \textuparrow & OEA \textuparrow  &  & \\
\midrule
\multicolumn{6}{l}{\textit{\textbf{Loss Function}} (fixed $\lambda=2.0$)} \\
Focal Loss & 58.1 & 63.2 & 49.8 & 77.4\% & 0.52s \\
\rowcolor{navyblue!10}
Standard CE & \textbf{59.8} & \textbf{65.4} & \textbf{51.4} & \textbf{86.7\%} & \textbf{0.39s} \\
\midrule
\multicolumn{6}{l}{\textit{\textbf{Weighting Factor}} (using Standard CE Loss)} \\
$\lambda = 1.0$ & 59.6 & 65.0 & \textbf{53.3} & 80.3\% & 0.42s \\
\rowcolor{navyblue!10}
$\lambda = 2.0$ & \textbf{59.8} & \textbf{65.4} & 51.4 & 86.7\% & \textbf{0.39s} \\
$\lambda = 3.0$ & 57.0 & 63.8 & 50.2 & \textbf{89.2\%} & 0.40s \\
\hline
\end{tabular}
}
\end{table}

\begin{table}[t]
\centering
\caption{\textbf{Ablation study on the impact of 3D geometry prior modeling in the VSFI module.} We compare the full Stream3D-VLM model with a fine-tuned Qwen2.5-VL-7B baseline and with variants that remove individual components.}
\label{tab:ablation_VSFI}
\setlength{\tabcolsep}{6pt}
\resizebox{0.8\textwidth}{!}{%
\begin{tabular}{c|ccc|cc}
\toprule
\multirow{2}{*}{\textbf{Settings}} & \multicolumn{3}{c|}{\textbf{Stream3D-Bench}} & \multicolumn{2}{c}{\textbf{VSI-Bench}} \\
\cmidrule(lr){2-4} \cmidrule(lr){5-6}  
& NA \textuparrow & MCA \textuparrow & OEA \textuparrow & NA \textuparrow & MCA \textuparrow \\
\midrule
Baseline (Visual Only) & 46.0 & 52.9 & 44.6 & 42.9 & 46.2 \\
\midrule
\textit{w/o} Camera Tokens & 55.4 & 60.4 & 46.3 & 60.5 & 61.4 \\
\textit{w/o} Geometry Tokens & 52.9 & 58.6 & 50.2 & 56.1 & 55.7 \\
\midrule
Fusion: Addition & 57.6 & \textbf{65.8} & 49.2 & 63.5 & 63.4 \\
Fusion: Concat. + MLP & 53.5 & 60.3 & 50.1 & 60.8 & 62.1 \\
\rowcolor{navyblue!10}
\textbf{Full Model (Cross-Attn.)} & \textbf{59.8} & 65.4 & \textbf{51.4} & \textbf{65.7} & \textbf{66.0} \\
\hline
\end{tabular}
}
\end{table}

\begin{table}[t]
\centering
\caption{\textbf{Comparison of token compression strategies at a 50\% retention ratio.} Unlike baseline methods that rely solely on semantic redundancy or attention scores, GAVC performs spatially guided dynamic updates, thereby preserving geometric consistency under streaming video inputs.}
\label{tab:compare_token-compression}
\resizebox{0.8\textwidth}{!}{%
\begin{tabular}{c|ccc|cccc}
\toprule
\multirow{2}{*}{\textbf{Methods}} & \multicolumn{3}{c|}{\textbf{Stream3D-Bench}} & \multicolumn{4}{c}{\textbf{ScanQA}} \\
\cmidrule(lr){2-4} \cmidrule(lr){5-8}  
& NA \textuparrow & MCA \textuparrow & OEA \textuparrow & B-4 \textuparrow & ROUGE \textuparrow & CIDEr \textuparrow & EM \textuparrow \\
\midrule
\multicolumn{8}{l}{\textit{\textbf{Geometry-Unaware Baselines}}} \\ 
Random & 35.6 & 40.2 & 36.8 & 8.5 & 32.6 & 72.4 & 18.8 \\
Avg. Pooling & 47.8 & 52.9 & 45.2 & 11.4 & 41.6 & 94.5 & 24.0\\
VisionZip & 49.2 & 53.8 & 41.6 & 13.0 & 43.8 & 87.2 & 24.4 \\
\midrule
\multicolumn{8}{l}{\textit{\textbf{Geometry-Adaptive Compression}}} \\ 
\rowcolor{navyblue!10}
\textbf{GAVC (Ours)} & \textbf{59.8} & \textbf{65.4} & \textbf{51.4} & \textbf{17.8} & \textbf{50.2} & \textbf{104.5} & \textbf{30.9} \\
\hline
\end{tabular}
}
\end{table}

\begin{figure}[t]
  \centering
  \includegraphics[width=\linewidth]{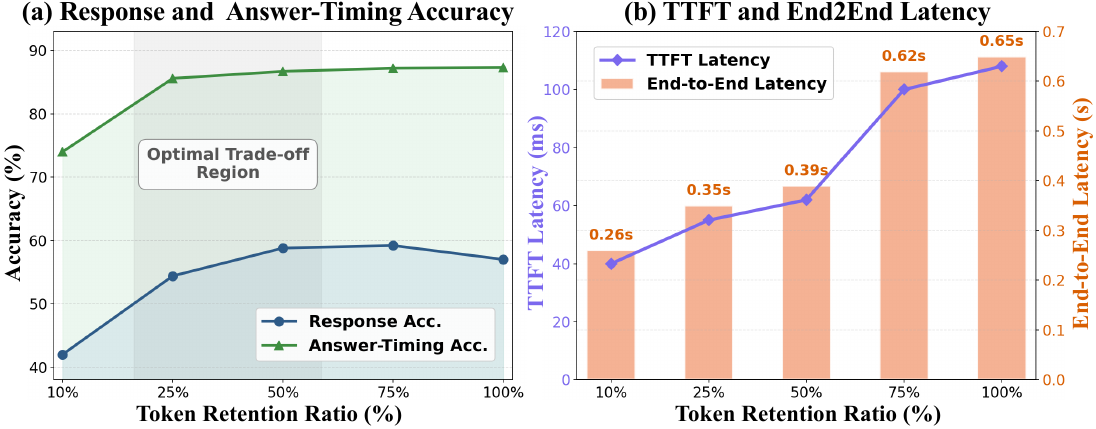}
  \caption{\textbf{Ablation study of the token retention ratio in the GAVC module.} We analyze the trade-off between performance and efficiency by reporting response accuracy, answer-timing accuracy, TTFT, and end-to-end latency on Stream3D-Bench. }
  \label{fig:ablation_GAVC}
\end{figure}

\subsection{Ablation Studies}
\textbf{Ablation Analysis of Streaming Loss Configurations.} We explore different loss functions and weighting factors for streaming prediction to balance performance and efficiency. Table~\ref{tab:ablation_streamingloss} shows that standard cross-entropy and focal loss perform similarly. However, over-weighting the streaming loss harms overall performance, while underweighting reduces response time accuracy. Optimal results occur at a streaming-to-generation loss weight ratio of 2.0.

\noindent\textbf{Effects of Visual-Spatial Feature Integration (VSFI).} Table~\ref{tab:ablation_VSFI} evaluates strategies for combining latent geometric features. Using either camera information or geometry tokens alone improves over the baseline, while combining both performs best. Moreover, a fusion design with stacked cross-attention blocks and skip connections outperforms patch-level addition or MLP-based concatenation.

\noindent\textbf{Impacts of Geometry-Adaptive Voxel Compression (GAVC).} Table~\ref{tab:compare_token-compression} compares GAVC with other token compression baselines such as random pruning, average pooling, and VisionZip~\cite{yang2025visionzip}. Unlike methods based on semantic redundancy or attention scores that ignore 3D structure, GAVC dynamically updates voxels guided by spatial properties, better handling irregular voxel distributions. As shown in Figure~\ref{fig:ablation_GAVC}, GAVC maintains competitive accuracy with substantially reduced latency, even at a 25\% token retention ratio.

\section{Conclusion}
In this paper, we propose Stream3D-VLM, the first online 3D spatial understanding model solely on streaming video. We reformulate streaming control as a next-token prediction problem, enabling the model to learn when to respond or keep silent. For continuous 3D scene comprehension, we introduce a VSFI module that incrementally injects temporally aligned geometric priors into the visual stream. We further propose a GAVC module to dynamically compress visual tokens guided by 3D structure, reducing long-context redundancy during online inference. Extensive experiments demonstrate that our model achieves leading performance across diverse online and offline 3D-language tasks.

%
%
\newpage
\bibliographystyle{splncs04}
\bibliography{main}

\clearpage
\appendix

\renewcommand{\labelitemi}{$\bullet$} %

\setcounter{section}{0}
\renewcommand{\thesection}{\Alph{section}}
\renewcommand{\theHsection}{\Alph{section}}

\setcounter{table}{0}
\renewcommand{\thetable}{A\arabic{table}}
\makeatletter
\renewcommand{\fnum@table}{Tab-\thetable}
\makeatother

\setcounter{figure}{0}
\renewcommand{\thefigure}{A\arabic{figure}}
\makeatletter
\renewcommand{\fnum@figure}{Fig-\thefigure}
\makeatother

\begin{center}
{\LARGE\bfseries Supplementary Material}
\end{center}
\vspace{10pt}

\noindent In this part, we provide more details and additional experimental results on our approach. The supplementary material is organized as follows:
\begin{itemize}
    \item \S~\ref{sec:metadata_computing_details}: Metadata computing details;
    \item \S~\ref{sec:gen_pipeline_details}: Detailed data generation pipeline;
    \item \S~\ref{sec:dataset_stats}: Stream3D-1M dataset statistics;
    \item \S~\ref{sec:bench_details}: Stream3D-Bench details;
    \item \S~\ref{sec:visualization}: More visualization results;
    \item \S~\ref{sec:evaluate_prompts}: Evaluation prompts for offline models.
\end{itemize}

\section{Metadata Computing Details}
\label{sec:metadata_computing_details}
This section details the computational logic for the metadata introduced in Section 3.2 of the main paper.

\subsection{Visibility and Occlusion Reasoning}
To determine the visibility of an object instance $O_i$ at frame $t$, we project its 3D mesh vertices $V_i$ onto the 2D image plane via the camera intrinsic matrix $K$ and the extrinsic pose $E_t$. Let $\mathbf{p}_{proj} = (u, v)$ denote the projected pixel coordinates of a vertex $\mathbf{x} \in V_i$, and $d_{\mathbf{x}}$ be its corresponding depth in the camera coordinate system. We determine the vertex-level visibility $\mathcal{V}_t(\mathbf{x})$ by enforcing geometric consistency between the projected vertex depth and the sensor depth map $D_t(u, v)$:

\begin{equation}
    \mathcal{V}_t(\mathbf{x}) = \mathbb{I} \left( |d_{\mathbf{x}} - D_t(u, v)| < \tau_{occ} \right) \cdot \mathbb{I} \left( 0 \le u < W, 0 \le v < H \right),
\end{equation}
where $\mathbb{I}(\cdot)$ represents the indicator function, and $\tau_{occ}$ denotes the occlusion tolerance threshold (empirically set to 0.05m).

We quantify the object visibility score for the current frame as the ratio of the number of currently visible vertices to the maximum number of visible vertices observed across the entire video sequence, inspired by recent work~\cite{zhang2025spar,wang2025n3d}. Finally, to obtain a binary visibility label, we threshold this ratio using values adapted to specific tasks.

\subsection{Camera Kinematics Computation}
We analyze the geometric properties of the camera pose trajectory, denoted as $\{ \mathbf{P}_t \}_{t=0}^{T}$. Let $\mathbf{t}_t \in \mathbb{R}^3$ represent the camera center (translation) and $\mathbf{R}_t \in SO(3)$ represent the orientation at frame $t$. We compute the following kinematic metrics:

\begin{itemize}
    \item \textbf{Path Length:} Calculated as the cumulative Euclidean distance traversed by the camera center across consecutive frames: 
    \begin{equation}
    L = \sum_{i=1}^{T} \| \mathbf{t}_i - \mathbf{t}_{i-1} \|_2.
    \end{equation}
    
    \item \textbf{Displacement:} Defined as the Euclidean distance between the initial and final camera positions: 
    \begin{equation}
    \Delta = \| \mathbf{t}_T - \mathbf{t}_0 \|_2.
    \end{equation}
    
    \item \textbf{Direction:} Defined as the azimuthal angle (\emph{i.e.}, clock position) of the final position relative to the initial pose. This is computed by projecting the displacement vector $(\mathbf{t}_T - \mathbf{t}_0)$ onto the local coordinate system of the initial camera $\mathbf{P}_0$.
    
    \item \textbf{Rotation:} Defined as the variation in horizontal orientation between the start and end frames. It is obtained by calculating the signed angle between the optical axes (forward vectors) of $\mathbf{R}_T$ and $\mathbf{R}_0$ projected onto the horizontal plane.
\end{itemize}

\subsection{Geometric Measurement}
Leveraging the dense point clouds derived from 3D instance segmentation, we compute the following metrics to characterize the geometric properties of the scene and objects:

\begin{itemize}
    \item \textbf{Object-Camera Relationship:} We analyze the egocentric spatial relation of an object with respect to the camera. Specifically, the \textit{Direction} (discretized as clock positions) is computed based on the object centroid in the camera's local coordinate system. The \textit{Distance} is defined as the minimum Euclidean distance between the camera center and the set of points belonging to the object instance.
    
    \item \textbf{Inter-Object Distance:} To quantify the spatial separation between two objects, we compute the minimum Euclidean distance between their respective 3D point sets. This calculation is efficiently accelerated using a KD-Tree structure to query the nearest neighbor points.
    
    \item \textbf{Object Size:} We estimate the spatial extent of an object by constructing an Oriented Bounding Box (OBB) around its point cloud. The dimensions are parameterized by the length, width, and height of this bounding volume.
    
    \item \textbf{Room Area:} The approximate area of the room is derived from the global scene point cloud. We project the scene's bounding box onto the horizontal plane and calculate the product of its length and width.
\end{itemize}

\section{Detailed Data Generation Pipeline}
\label{sec:gen_pipeline_details}

\subsection{Rule-based Generation Templates}
We constructed diverse linguistic templates for each rule-based task.
\begin{itemize}
    \item \textbf{Ego-Motion Estimation:} Questions focus on monitoring future motion (Forward) or recalling past motion (Backward). 
    \textit{Example:} ``How much total distance has the camera traveled in the past $\{N\}$ seconds?''
    
    \item \textbf{Object--Camera Relationship:} Covers Distance, Direction (\emph{e.g.}, ``to your left"), and Location (Clock position + Distance).
    \textit{Example:} ``Where is the $\{Object\}$ relative to me right now?'' (Realtime) vs. ``I can't see the $\{Object\}$ anymore. Where is it relative to me?'' (Backward).
    
    \item \textbf{Environment Measurement:}
    Includes Object Size, Room Area, and Inter-object Distance. 
    \textit{Example:} What is the shortest distance between the $\{Object\_A\}$ and the $\{Object\_B\}$ in meters?
    
    \item \textbf{Object Chronology:}
    Involves counting unique instances, determining appearance order, and timestamps.
    \textit{Example:} ``How many $\{Object\_Class\}$ have you seen so far?'' or ``Did you see the $\{Object\_A\}$ before or after the $\{Object\_B\}$?''
\end{itemize}
Table~\ref{tab:question_templates} presents the complete collection of question templates for all 29 tasks, hierarchically structured according to three temporal interaction modes and five cognitive categories.

\definecolor{highlight}{RGB}{200, 0, 0} 
\newcommand{\hl}[1]{\textcolor{highlight}{#1}}
\begin{table*}[h!]
    \renewcommand{\tabularxcolumn}[1]{m{#1}} 
    \centering
    \tiny
    \renewcommand{\arraystretch}{1.1} 
    \setlength{\tabcolsep}{3pt}       
    \caption{\textbf{Taxonomy and question templates of Stream3D-Bench.} The benchmark comprises 29 tasks, hierarchically structured according to temporal interaction modes and cognitive categories. \hl{Highlighted} variables are dynamically instantiated for each scene.}
    \label{tab:question_templates}
    
    \begin{tabularx}{\textwidth}{c|c|l|X}
        \toprule
        \textbf{Mode} & \textbf{Category} & \textbf{Specific Task} & \textbf{Question Template / Example} \\
        \midrule
        
        \multirow{21}{*}{\rotatebox{90}{\textbf{Backward Tracing}}} 
        & \multirow{9}{*}[0.5em]{\textbf{Ego-Motion}}
          & Cam. Direction & \textit{Where is the current camera location relative to its position \hl{\{n\}} seconds ago?} \\ 
        & & Cam. Path & \textit{What is the total path length covered by the camera during the last \hl{\{n\}} seconds?} \\
        & & Cam. Rotation & \textit{What was the camera's horizontal rotation angle over the past \hl{\{n\}} seconds?} \\
        & & Cam. Displacement & \textit{How far has the camera moved during the last \hl{\{n\}} seconds?} \\
        & & Cam. Comprehensive & \textit{Provide a detailed summary of the camera's motion over the last \hl{\{n\}} seconds.} \\
        \cmidrule{2-4}
        
        & \textbf{Env. Measure.} 
          & Room Area & \textit{Based on the observed footage, estimate the room's length, width, and area.} \\
        \cmidrule{2-4}
        
        & \multirow{6}{*}[0.5em]{\textbf{Obj-Cam Rel.}} 
          & Obj-Cam Distance & \textit{The \hl{\{object\}} is currently out of view. How far away is it from me?} \\
        & & Obj-Cam Direction & \textit{Where is the \hl{\{object\}} relative to me now? Please use directional terms.} \\
        & & Obj-Cam Location & \textit{Provide both the clock direction and the distance of the \hl{\{object\}} relative to me.} \\
        \cmidrule{2-4}
        
        & \multirow{4}{*}{\textbf{Chronology}} 
          & Object Counting & \textit{How many \hl{\{category\}}(s) have you seen so far?} \\
        & & Appearance Time & \textit{At what timestamp (in seconds) did the \hl{\{object\}} first appear?} \\
        & & Appearance Order & \textit{What is the first-time appearance order of \hl{\{choice a\}}, \hl{\{choice b\}}, \hl{\{choice c\}}?} \\
        
        \midrule
        
        \multirow{10}{*}[-0.5em]{\rotatebox{90}{\textbf{Realtime Perception}}} 
        & \multirow{3}{*}{\textbf{Env. Measure.}} 
          & Object Size & \textit{Estimate the size of the \hl{\{object\}}.} \\
        & & Obj-Obj Distance & \textit{What is the distance between \hl{\{object A\}} and \hl{\{object B\}} currently?} \\
        \cmidrule{2-4}
        
        & \multirow{2}{*}{\textbf{Obj-Cam Rel.}} 
          & Absolute Distance & \textit{How far is this \hl{\{object\}} from my current position?} \\
        & & Relative Distance & \textit{Which of the following objects (\hl{\{a, b, c, d\}}) is closest to me?} \\
        \cmidrule{2-4}
        
        & \multirow{5}{*}[0.5em]{\textbf{Attributes}} 
          & Object Property & \textit{What is the color/material/shape of the \hl{\{object\}} currently in view?} \\
        & & Object Position & \textit{Is the \hl{\{object\}} currently located on the floor or on another object?} \\
        & & Object Recognition & \textit{Identify the \hl{\{object\}} currently at the center of the view.} \\
        
        \midrule
        
        \multirow{17}{*}{\rotatebox{90}{\textbf{Forward Response}}} 
        & \multirow{9}{*}{\textbf{Ego-Motion}} 
          & Cam. Direction & \textit{Where will the camera be located compared to here in \hl{\{n\}} seconds?} \\
        & & Cam. Path & \textit{In \hl{\{n\}} seconds, how much ground will the camera cover in total?} \\
        & & Cam. Rotation & \textit{How many degrees will the camera turn horizontally in \hl{\{n\}} seconds?} \\
        & & Cam. Displacement. & \textit{How far will the camera be from its current position in \hl{\{n\}} seconds?} \\
        & & Cam. Comprehensive & \textit{Give a comprehensive report on the camera's motion over the next \hl{\{n\}} seconds.} \\
        \cmidrule{2-4}
        
        & \textbf{Env. Measure.} 
          & Room Area & \textit{Once the room is sufficiently covered, tell me its length, width, and area.} \\
        \cmidrule{2-4}
        
        & \multirow{6}{*}{\textbf{Obj-Cam Rel.}} 
          & Obj-Cam Distance & \textit{Wait \hl{\{n\}} seconds, then tell me the distance to the \hl{\{object\}}.} \\
        & & Obj-Cam Location & \textit{In \hl{\{n\}} seconds, report both the clock direction and distance of the \hl{\{object\}} relative to me.
        } \\
        & & Obj-Cam Direction & \textit{Wait \hl{\{n\}} seconds, then indicate the direction of the \hl{\{object\}}.} \\
        & & Object Finding & \textit{I'm moving, help locate the \hl{{\{object\}}} and report its clock direction and distance.} \\
        
        \bottomrule
    \end{tabularx}
\end{table*}

\subsection{QA Transfer with VLM Verification}
For semantic tasks (Attribute, Position, Recognition), we adapt static ScanQA data.
\begin{enumerate}
    \item \textbf{Temporal Grounding:} We align the static question to the video timeline by identifying the frame $t^*$ where the target object is most clearly visible (highest visibility ratio).
    \item \textbf{VLM Verification:} We employ GPT-5~\cite{openai_gpt-5} with the prompt: \textit{``You are an expert in visual question answering verification. Please examine whether the provided image frame contains sufficient visual evidence to support the given Question and Answer pair. You must assess if the target object is clearly perceivable. Return `1` if the answer can be strictly inferred from the visual content, and `0` otherwise.''} to ensure the question is answerable from the visual stream alone, filtering out questions relying on unobservable context.
\end{enumerate}

\subsection{Construction of Target Object Whitelist}
\begin{figure*}[h]
  \centering
  \includegraphics[width=\linewidth]{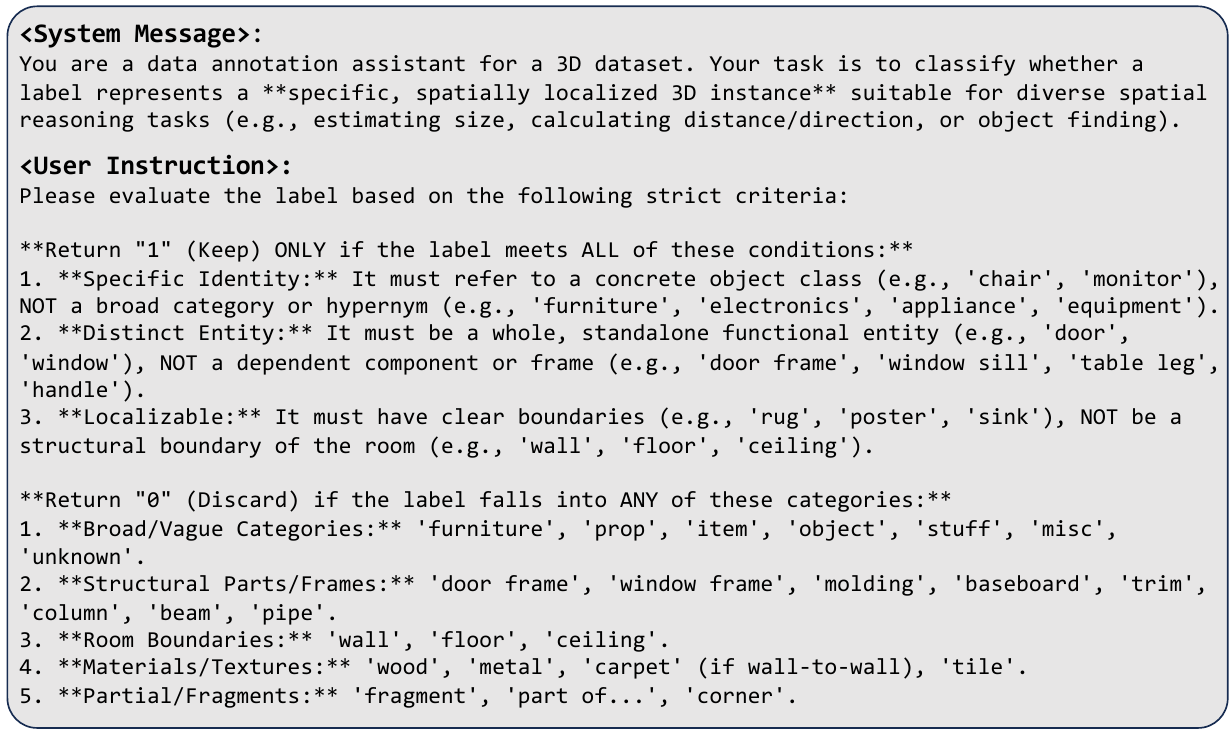}
  \caption{Prompts employed to curate the Target Object Whitelist for object-centric 3D online QA.}
  \label{fig:object_filtering}
\end{figure*}

When constructing streaming 3D QA pairs for object-centric tasks, we observed that many object labels are noisy, ambiguous, or otherwise unsuitable for meaningful question generation. To ensure high-quality QA, we employed GPT-5~\cite{openai_gpt-5} to rigorously curate a Target Object Whitelist from the full label sets of ScanNet and ScanNet++. We adopted a unified high-precision filtering strategy based on three core criteria:
\begin{enumerate}
    \item \textbf{Specific Identity:} Retain concrete object classes (\emph{e.g.}, chair, monitor) while discarding broad hypernyms (\emph{e.g.}, furniture, electronics) to ensure semantic precision.
    \item \textbf{Distinct Entity:} Select standalone functional entities rather than dependent parts (\emph{e.g.}, retain door but discard door frame).
    \item \textbf{Spatial Localizability:} Prioritize objects with finite, well-defined boundaries, excluding continuous structural elements such as floor, ceiling, and wall.
\end{enumerate}
The prompts used to curate the Target Object Whitelist are shown in Figure~\ref{fig:object_filtering}.

\noindent\textbf{Note:} While structural elements like walls can theoretically support specific tasks (\emph{e.g.}, Object-to-Object Distance), they are unsuitable for the majority of object-centric tasks (\emph{e.g.}, Object Size or Object Counting). To maintain consistency across the benchmark, we applied this whitelist globally. This approach ensures a clear and consistent definition of ``object" across tasks, prevents referential ambiguity (\emph{e.g.}, distinguishing between multiple walls in a single view), and focuses evaluation on the model’s ability to perceive distinct 3D instances.

\begin{figure*}[t]
  \centering
  \includegraphics[width=\linewidth]{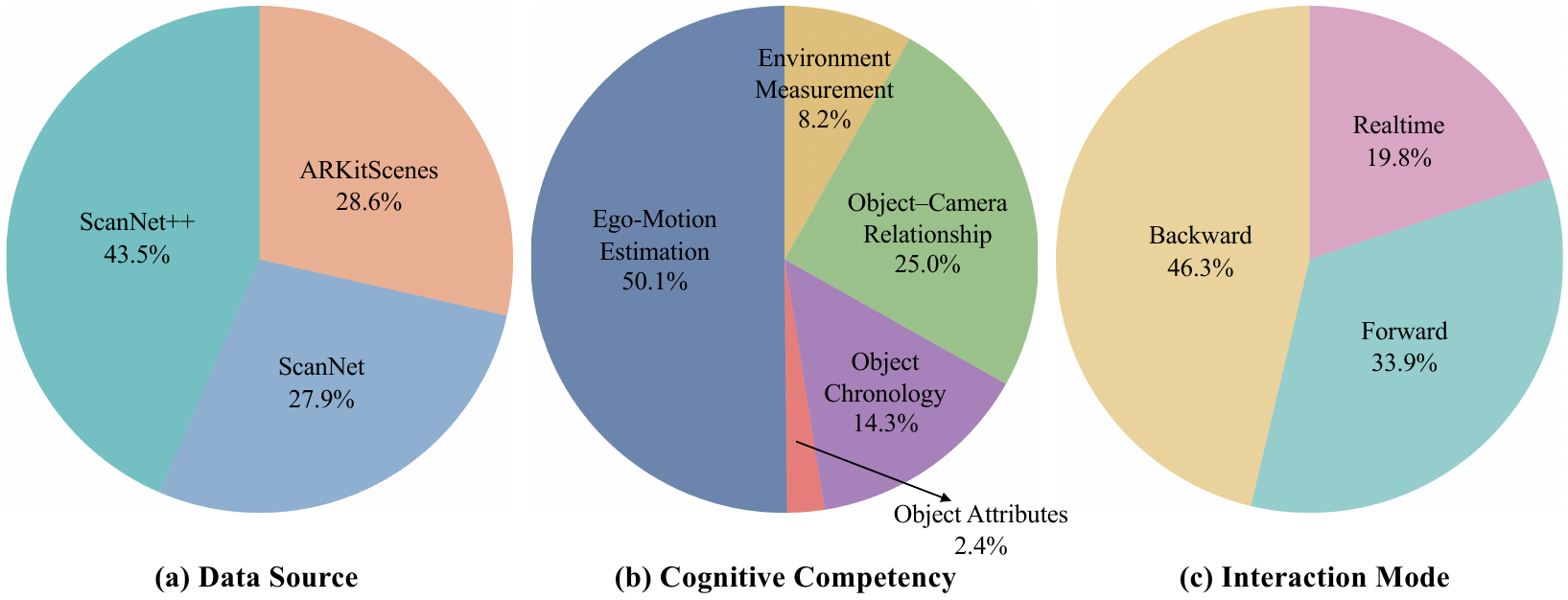}
  \caption{\textbf{Statistical distribution of Stream3D-1M dataset.} 
  The figure visualizes the composition of the training set (over 1M samples) across three key dimensions: 
  \textbf{(a) Data Source}, highlighting that ScanNet++ contributes the plurality of QA pairs (43.5\%) due to its dense annotations; 
  \textbf{(b) Task Category}, dominated by Ego-Motion Estimation tasks (50.1\%) which serve as the foundation for spatial tracking; 
  and \textbf{(c) Interaction Mode}, demonstrating a strong emphasis on long-term memory (Backward, 46.3\%) and active monitoring (Forward, 33.9\%).}
  \label{fig:dataset_stats}
\end{figure*}

\section{Stream3D-1M Dataset Statistics}
\label{sec:dataset_stats}

We present a comprehensive statistical analysis of the \textbf{Stream3D-1M Dataset}, a large-scale dataset containing approximately \textbf{1 million} (1,003,203) question-answer pairs derived from \textbf{5,154} unique scans. This dataset is designed to train Multimodal Large Language Models (MLLMs) with robust streaming 3D capabilities. The detailed composition is summarized in Figure~\ref{fig:dataset_stats} and Table~\ref{tab:dataset_stats}.

\noindent\textbf{Scene Diversity and Data Sources.} 
To ensure the model generalizes across diverse environments, our dataset aggregates scenes from three distinct sources. As detailed in Table~\ref{tab:dataset_stats} (Top), \textbf{ARKitScenes}~\cite{baruch2021arkitscenes} contributes the majority of raw video data, accounting for \textbf{60.1\%} of the total videos. This vast diversity is crucial for preventing the model from overfitting to the specific interior styles of ScanNet-like environments. \textbf{ScanNet}~\cite{dai2017scannet}, and \textbf{ScanNet++}~\cite{yeshwanth2023scannet++} contribute 23.3\% and 16.6\% of the videos, respectively, serving as the foundation for fine-grained semantic and geometric reasoning.

\noindent\textbf{QA Distribution and Density.} 
While ARKitScenes offers scene breadth, \textbf{ScanNet++} provides annotation depth. Despite comprising fewer scenes, ScanNet++ contributes the largest portion of QA pairs (\textbf{43.5\%}), reflecting its high-quality, dense annotations that support complex spatial reasoning. Conversely, ARKitScenes contributes 28.6\% of the QAs, primarily focusing on camera motion and trajectories, balancing the dataset between semantic richness and kinematic diversity.

\noindent\textbf{Task Category Distribution.} 
The distribution of task types reflects the fundamental requirements of a streaming 3D assistant:
\begin{itemize}
    \item \textbf{Ego-Motion Estimation (50.1\%):} This category dominates the dataset. Since ego-motion estimation is a prerequisite for spatial tracking and is applicable to all data sources, it serves as a large-scale pre-training foundation.
    \item \textbf{Object-Camera Relation (25.0\%) \& Object Chronology (14.3\%):} These tasks constitute the core of spatial interaction, requiring the model to update object states relative to the viewer dynamically.
    \item \textbf{Environment Measurement (8.2\%) \& Object Attributes (2.4\%):} These tasks are inherently scarcer as they rely on specific geometric conditions or high-quality VLM-verified semantic attributes, ensuring high precision over quantity.
\end{itemize}

\noindent\textbf{Temporal Mode Distribution.} 
The dataset emphasizes memory and prediction capabilities. \textbf{Backward Tracing} accounts for \textbf{46.3\%} of the data, training the model's long-term history retention. \textbf{Realtime Perception} (19.8\%) and \textbf{Forward Response} (33.9\%) ensure the model remains responsive to current and future events.

\section{Stream3D-Bench Details}
\label{sec:bench_details}

\begin{figure*}[t]
  \centering
  \includegraphics[width=\linewidth]{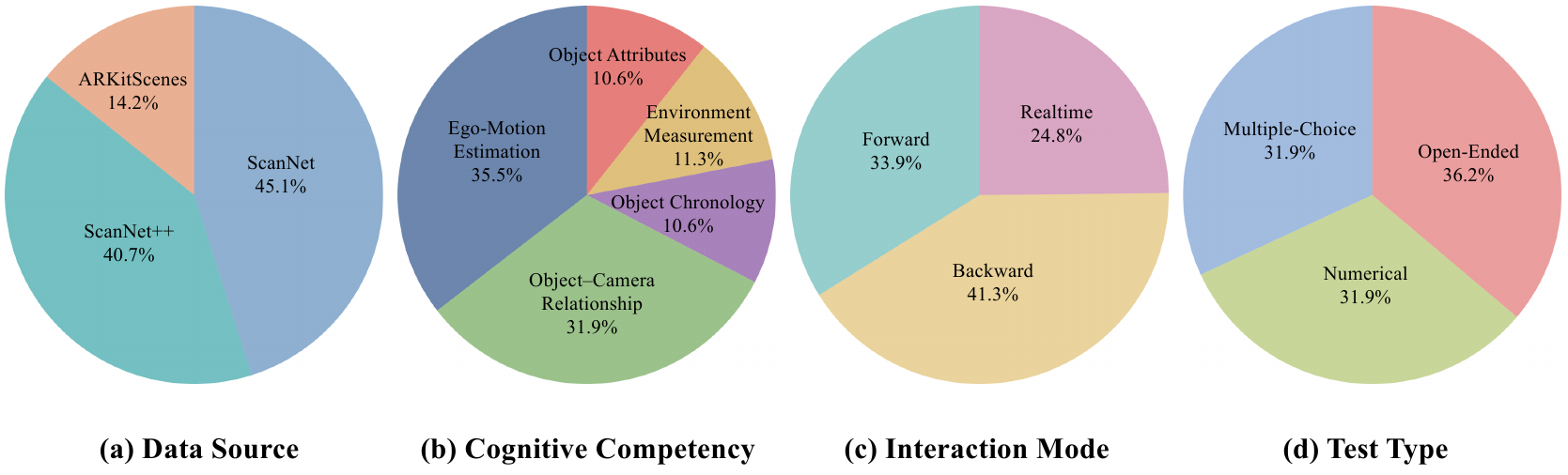}
  \caption{\textbf{Statistical distribution of Stream3D-Bench.} 
  The four pie charts illustrate the composition of the benchmark across diverse dimensions: 
  \textbf{(a) Data Source}, showing a robust integration of ScanNet, ScanNet++, and ARKitScenes; 
  \textbf{(b) Task Category}, covering five core cognitive capabilities; 
  \textbf{(c) Interaction Mode}, balancing tasks across Memory (Backward), Observation (Realtime), and Monitoring (Forward) phases; 
  and \textbf{(d) Answer Type}, maintaining a uniform distribution among Open-ended, Numerical, and Multiple-choice formats.}
  \label{fig:bench_stats}
\end{figure*}

To systematically evaluate the capabilities of MLLMs in online 3D spatial understanding, we construct \textbf{Stream3D-Bench}, a comprehensive benchmark comprising \textbf{10,037} high-quality samples derived from \textbf{518} unique videos. As shown in Figure~\ref{fig:bench_stats} and Table~\ref{tab:bench_stats}, the benchmark is rigorously balanced across data sources, task categories, interaction modes, and answer types, ensuring a fair and robust assessment.

\begin{table}[h]
\centering
\scriptsize
\begin{minipage}[t]{0.48\linewidth}
\centering
\caption{Detailed statistics of the Stream3D-1M dataset.}
\label{tab:dataset_stats}
\begin{tabular}{l|c|c}
\toprule
\textbf{Attribute} & \textbf{Count} & \textbf{Percent} \\
\midrule
\multicolumn{3}{l}{\textit{Scan Count by Source}} \\
ARKitScenes & 3,098 & 60.1\% \\
ScanNet & 1,201 & 23.3\% \\
ScanNet++ & 855 & 16.6\% \\
\textbf{Total Scans} & \textbf{5,154} & \textbf{100.0\%} \\
\midrule
\multicolumn{3}{l}{\textit{QA Count by Source}} \\
ScanNet++ & 436,812 & 43.5\% \\
ScanNet & 279,859 & 27.9\% \\
ARKitScenes & 286,532 & 28.6\% \\
\textbf{Total QAs} & \textbf{1,003,203} & \textbf{100.0\%} \\
\midrule
\multicolumn{3}{l}{\textit{Task Category}} \\
Ego-Motion Estimation & 503,072 & 50.1\% \\
Object-Camera Relationship & 250,650 & 25.0\% \\
Object Chronology & 143,475 & 14.3\% \\
Environment Measurement & 82,126 & 8.2\% \\
Object Attributes & 23,880 & 2.4\% \\
\midrule
\multicolumn{3}{l}{\textit{Interaction Mode}} \\
Backward & 464,241 & 46.3\% \\
Forward & 340,153 & 33.9\% \\
Realtime & 198,809 & 19.8\% \\
\bottomrule
\end{tabular}
\end{minipage}
\hfill
\begin{minipage}[t]{0.48\linewidth}
\centering
\caption{Detailed statistics of Stream3D-Bench composition.}
\label{tab:bench_stats}
\begin{tabular}{l|c|c}
\toprule
\textbf{Attribute} & \textbf{Count} & \textbf{Percent} \\
\midrule
\multicolumn{3}{l}{\textit{Data Source}} \\
ScanNet & 4,525 & 45.1\% \\
ScanNet++ & 4,082 & 40.7\% \\
ARKitScenes & 1,430 & 14.2\% \\
\midrule
\multicolumn{3}{l}{\textit{Task Category}} \\
Ego-Motion Estimation & 3,560 & 35.5\% \\
Object-Camera Relationship & 3,204 & 31.9\% \\
Environment Measurement & 1,137 & 11.3\% \\
Object Attributes & 1,068 & 10.6\% \\
Object Chronology & 1,068 & 10.6\% \\
\midrule
\multicolumn{3}{l}{\textit{Interaction Mode}} \\
Backward & 4,145 & 41.3\% \\
Forward & 3,400 & 33.9\% \\
Realtime & 2,492 & 24.8\% \\
\midrule
\multicolumn{3}{l}{\textit{Test Type}} \\
Open-Ended & 3,629 & 36.2\% \\
Numerical & 3,204 & 31.9\% \\
Multiple-Choice & 3,204 & 31.9\% \\
\bottomrule
\end{tabular}
\end{minipage}
\end{table}

\noindent\textbf{Data Sources and Diversity.} 
The benchmark integrates diverse indoor 3D data to prevent overfitting to specific sensor characteristics. 
 Specifically, the data is sourced from \textbf{ScanNet} (312 scans), \textbf{ScanNet++} (100 scans), and \textbf{ARKitScenes} (106 scans). 
While ScanNet and ScanNet++ provide dense semantic annotations for object-centric tasks, ARKitScenes contributes significantly to camera motion tasks, enriching the diversity of trajectory dynamics.

\noindent\textbf{Task Categories.} 
Stream3D-Bench covers five core capability dimensions. To ensure balanced evaluation, we apply a stratified sampling strategy where most specific sub-tasks (\textit{Camera Path (Backward)}, \textit{Object Finding (Forward)}) are explicitly balanced to approximately \textbf{356} samples each. 
Exceptions occur only in tasks with strict geometric constraints, such as \textit{Room Area}, where valid samples are inherently scarcer.
The overall distribution across categories is:
\begin{itemize}
    \item \textbf{Ego-Motion Estimation (35.5\%):} The largest category, assessing ego-motion perception across all three datasets.
    \item \textbf{Object-Camera Relationship (31.9\%):} Evaluates spatial reasoning capabilities like distance and direction estimation.
    \item \textbf{Environment Measurement (11.3\%):} Focuses on quantifying scene attributes (\emph{e.g.}, object size, room area).
    \item \textbf{Object Attributes (10.6\%):} Tests attribute recognition and property understanding.
    \item \textbf{Object Chronology (10.6\%):} Assesses temporal awareness, such as appearance order and counting.
\end{itemize}

\noindent\textbf{Temporal Interaction Modes.} 
The benchmark evaluates agents across three distinct temporal phases, ensuring the model can handle memory, observation, and prediction:
\begin{itemize}
    \item \textbf{Backward Tracing (41.3\%):} Assessing long-term memory and history retrieval.
    \item \textbf{Realtime Perception (24.8\%):} Testing instant spatial perception and current-frame analysis.
    \item \textbf{Forward Response (33.9\%):} Evaluating future monitoring and asynchronous response capabilities.
\end{itemize}

\noindent\textbf{Answer Types.} 
To validate different output modalities, the benchmark maintains a near-perfect balance among answer formats: \textbf{Open-ended} (36.2\%), \textbf{Numerical} (31.9\%), and \textbf{Multiple-choice} (31.9\%). This uniform distribution prevents bias towards any specific answering style.

\clearpage
\section{More Visualization Results}
\label{sec:visualization}
\begin{figure*}[h]
  \centering
  \includegraphics[width=\linewidth]{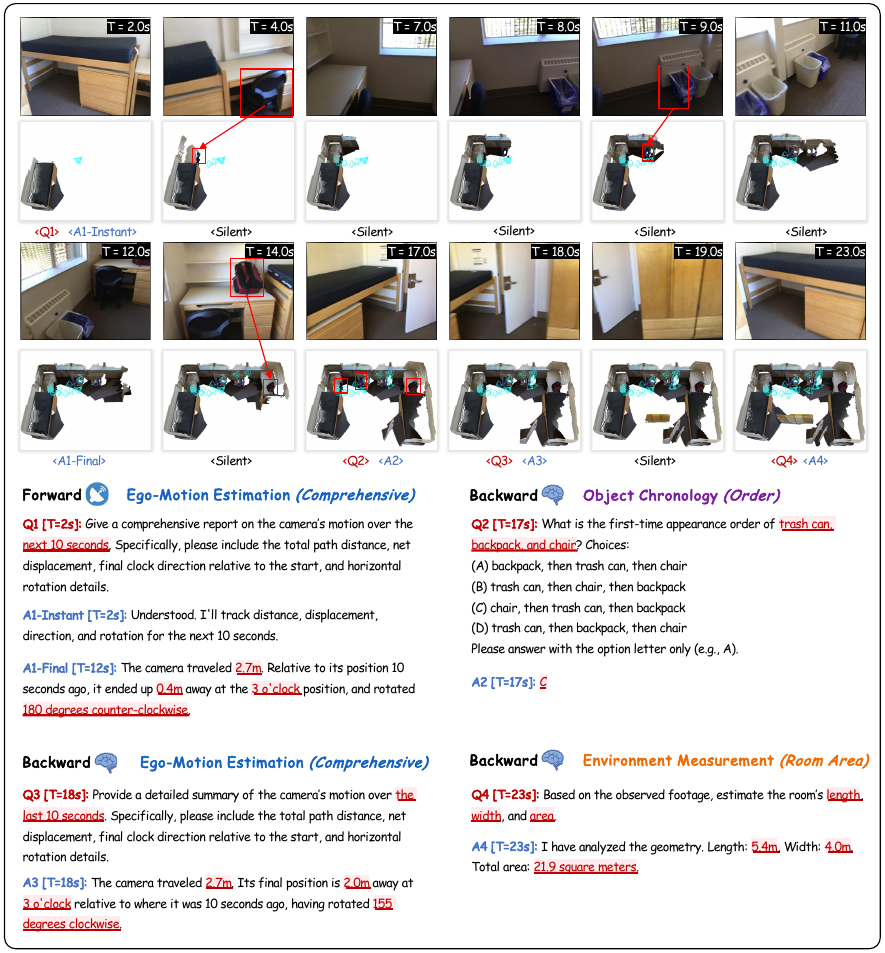}
  \caption{\textbf{Stream3D-Bench Examples (Part 1).} 
  }
  \label{fig:benchmark_examples_1}
\end{figure*}

\begin{figure*}[h]
  \centering
  \includegraphics[width=\linewidth]{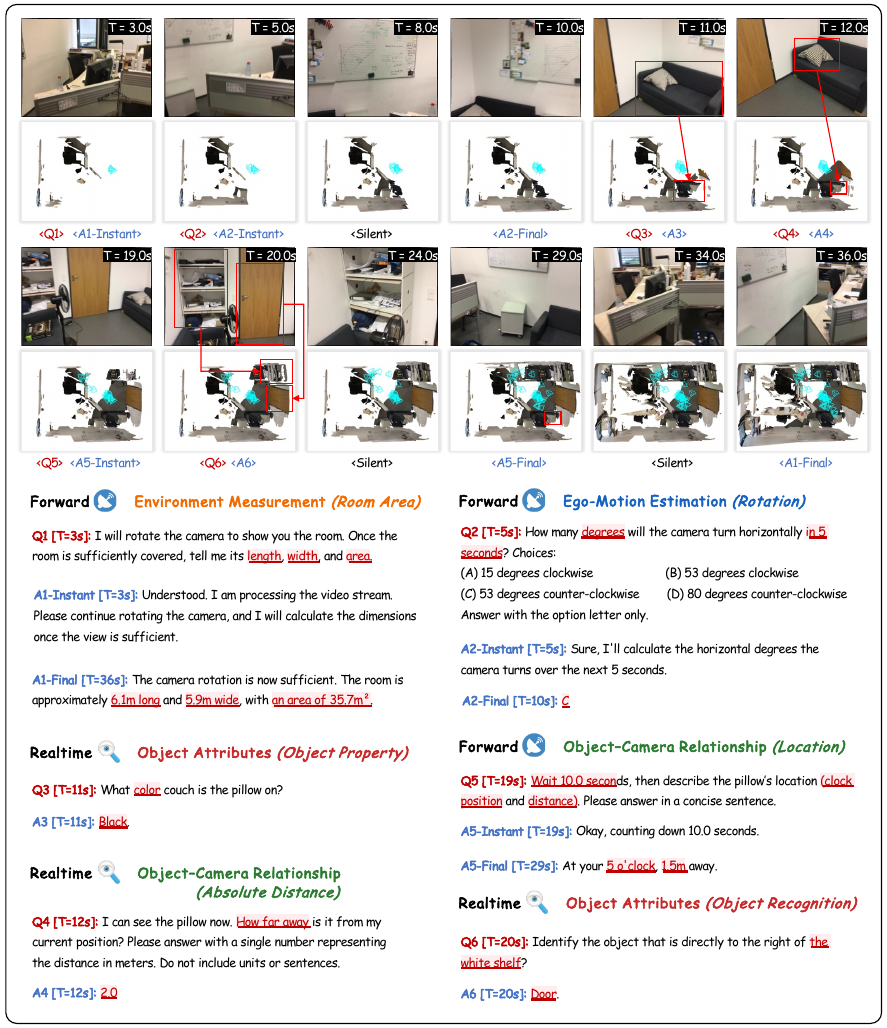}
  \caption{\textbf{Stream3D-Bench Examples (Part 2).} 
  }
  \label{fig:benchmark_examples_2}
\end{figure*}

\begin{figure*}[h]
  \centering
  \includegraphics[width=\linewidth]{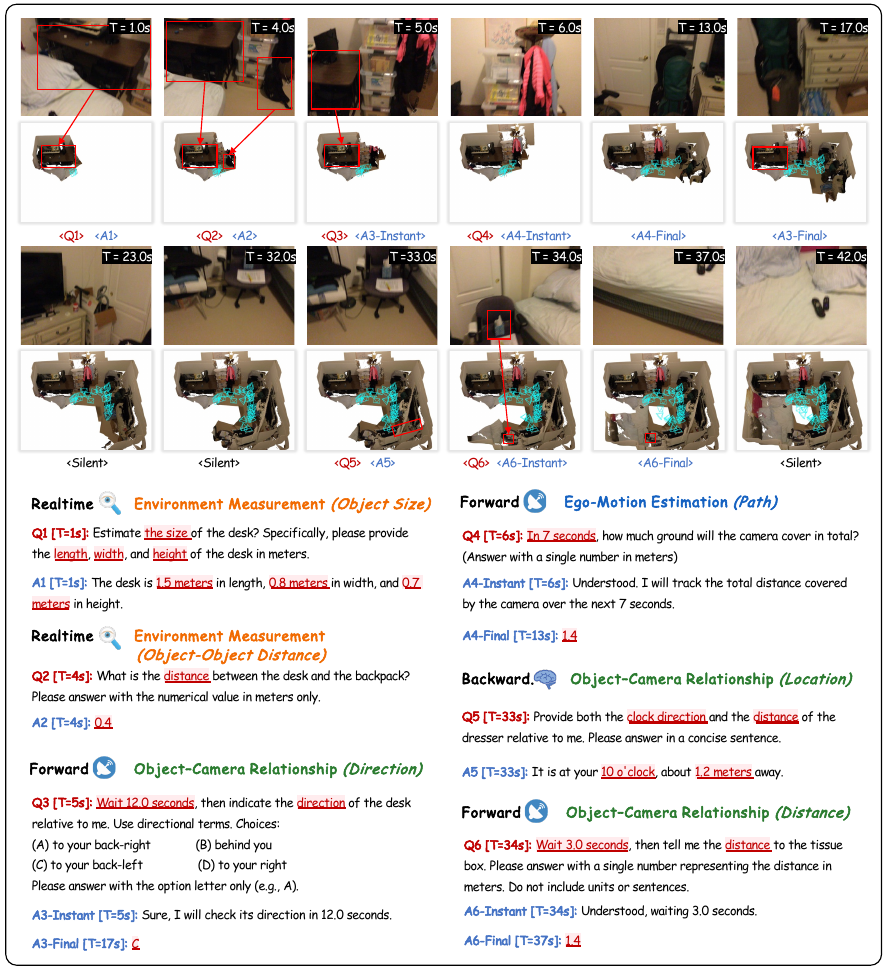}
  \caption{\textbf{Stream3D-Bench Examples (Part 3).} 
  }
  \label{fig:benchmark_examples_3}
\end{figure*}

\begin{figure*}[h]
  \centering
  \includegraphics[width=\linewidth]{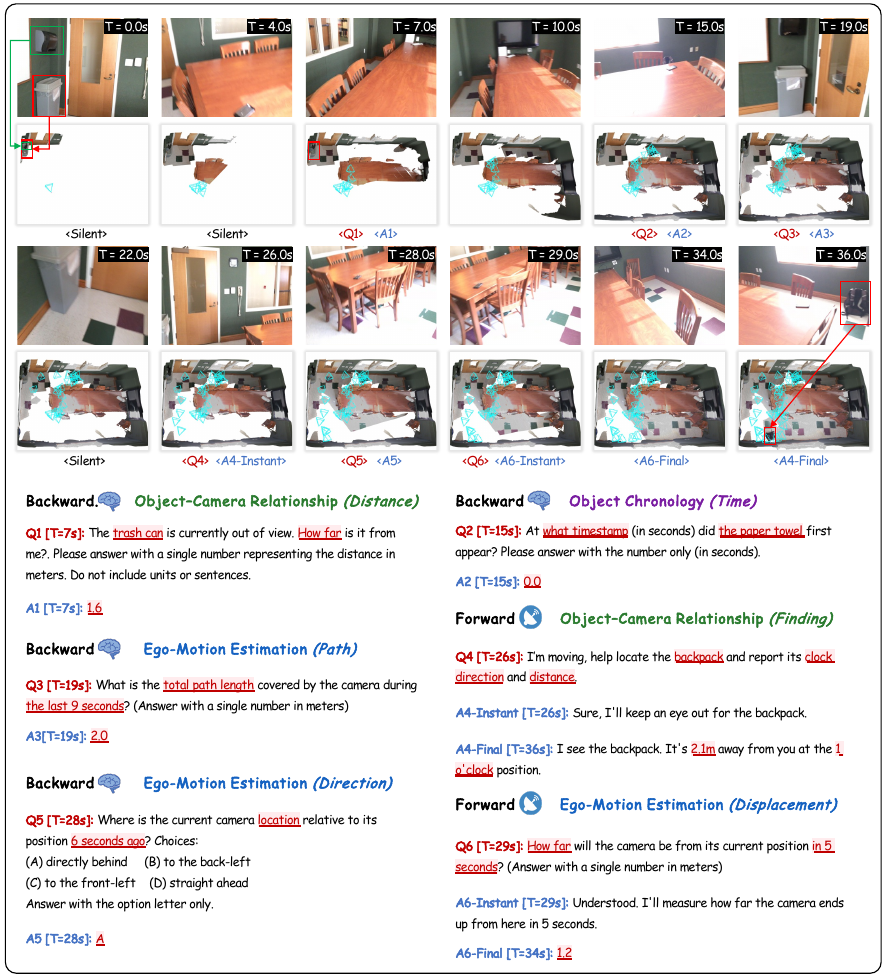}
  \caption{\textbf{Stream3D-Bench Examples (Part 4).} 
  }
  \label{fig:benchmark_examples_4}
\end{figure*}

\begin{figure*}[h]
  \centering
  \includegraphics[width=\linewidth]{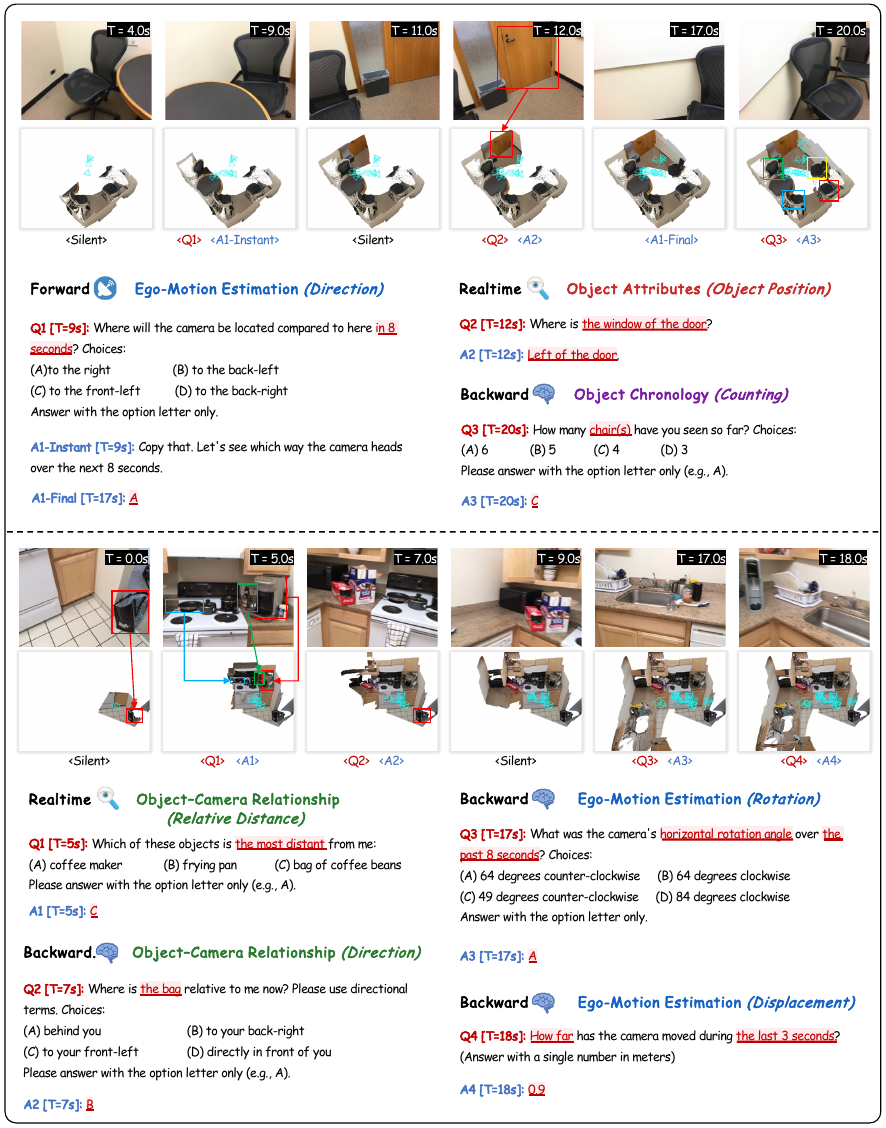}
  \caption{\textbf{Stream3D-Bench Examples (Part 5).} 
  }
  \label{fig:benchmark_examples_5}
\end{figure*}

In this section, we present comprehensive visualization examples covering all 29 tasks in Stream3D-Bench, hierarchically structured along three temporal interaction patterns and five cognitive categories, as illustrated in Figures~\ref{fig:benchmark_examples_1}–\ref{fig:benchmark_examples_5}.

\clearpage
\section{Evaluation Prompts for Offline Models}
\label{sec:evaluate_prompts}
\begin{figure*}[h]
  \centering
  \vspace{-5mm}
  \includegraphics[width=1.0\linewidth]{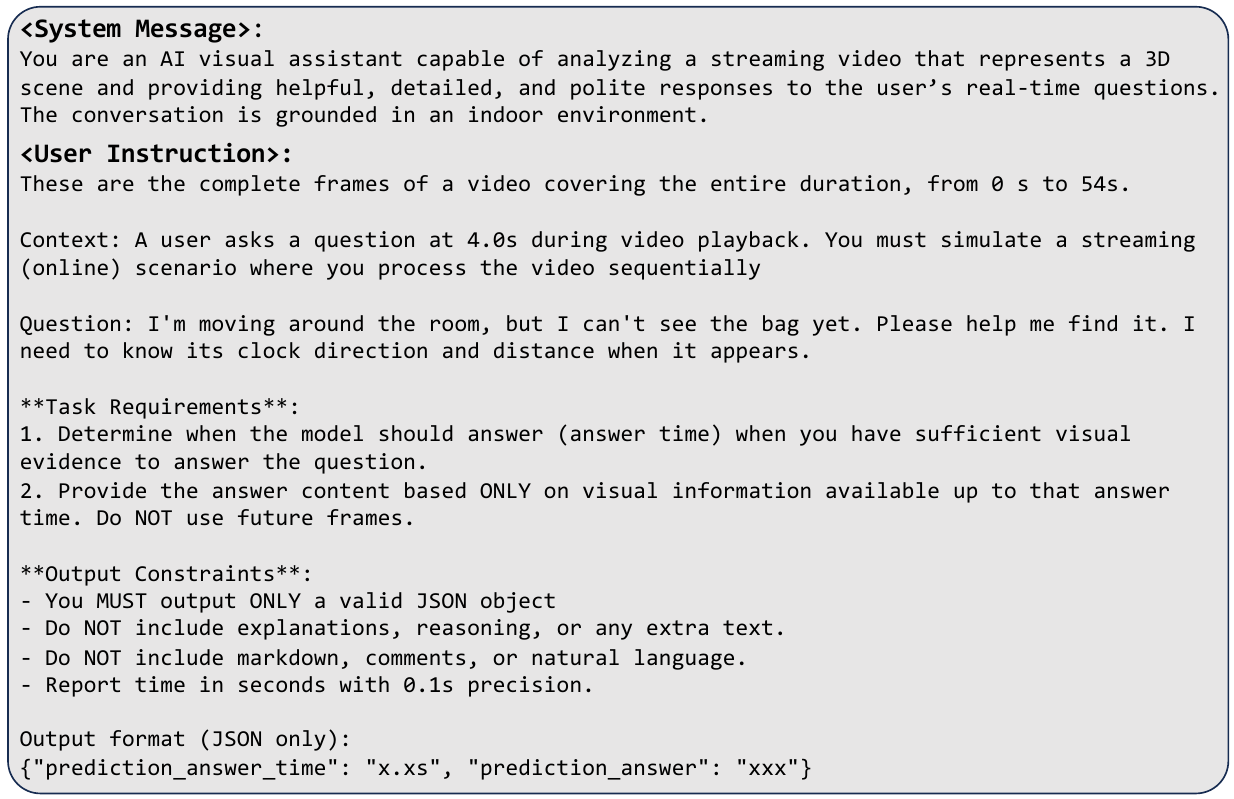}
  \vspace{-5mm}
  \caption{Evaluation prompts for offline MLLMs on our Stream3D-Bench.}
  \label{fig:evaluate_prompt}
  \vspace{-1mm}
\end{figure*}

Since existing offline models~\cite{hu2025g2,huang2025leo-vl,wang2025spatial3d,deng20253d-llava,kang2025robin3d,ouyang2025spacer} cannot natively process streaming videos or generate online responses, we adapt them to Stream3D-Bench by providing the full video as input and explicitly specifying the query time. The models are prompted to jointly predict when to respond and what answer to produce. As illustrated in Figure~\ref{fig:evaluate_prompt}, we explicitly inform the model of the total video duration and require it to simulate a streaming (online) scenario. Specifically, the query timestamp and the user question are given simultaneously, and the model must determine at what time after the query it has accumulated sufficient visual evidence to respond. The answer is constrained to rely only on visual information available up to the predicted response time.

Although this protocol enables a coarse evaluation of existing offline MLLMs on Stream3D-Bench, it inherently places them in a favorable and unrealistic setting, as the entire video is provided upfront, effectively granting access to more visual information than would be available in a true streaming scenario. Despite this disadvantageous comparison setup, experimental results consistently show that our real-time streaming 3D spatial understanding model significantly outperforms offline baselines across multiple evaluation metrics, including both answer accuracy and response-time precision. These results further demonstrate the superiority and effectiveness of our approach for online 3D spatial understanding. In future work, we plan to explore a broader range of streaming feed-forward 3D reconstruction models~\cite{lin2025depthv3,lan2025stream3r,yuan2026infinitevggt,su2026xstreamvggt,wang20254dvggt} to further investigate and enhance model generalization and stability~\cite{wei2023moire,wei2024physical}, thereby facilitating better adaptation to real-world embodied applications.

\end{document}